\documentclass{article}

  \usepackage[preprint]{neurips_2026}

\usepackage[utf8]{inputenc} 
\usepackage[T1]{fontenc}    

\usepackage{microtype}      

\usepackage{amsmath}        
\usepackage{amsfonts}       
\usepackage{amssymb}        
\usepackage{amsthm}         
\usepackage{mathtools}      
\usepackage{nicefrac}       
\usepackage{xfrac}          
\usepackage{booktabs}
\usepackage{multirow}
\usepackage{pifont}
\usepackage{microtype}
\usepackage{graphicx}
\usepackage{subcaption}
\usepackage{booktabs} 
\usepackage{multirow}
\usepackage{comment}
\usepackage{hyperref}
\usepackage{url}
\usepackage{wrapfig}
\usepackage{colortbl}
\usepackage{pifont}
\usepackage{xcolor}

\usepackage{colortbl}
\usepackage{siunitx}

\definecolor{bestcell}{HTML}{E8F5E9}
\definecolor{worstcell}{HTML}{FFEBEE}
\definecolor{headerblue}{HTML}{E3F2FD}
\usepackage{graphicx}       
\usepackage{tikz}           
\usepackage{subcaption}     
\usepackage{wrapfig}        
\usepackage{float}          
\usepackage{adjustbox}      

\usetikzlibrary{arrows.meta,backgrounds,fit,decorations.pathreplacing,patterns,shapes.geometric,3d}
\usepackage{tkz-euclide}    
\usetikzlibrary{shapes.geometric}
\usetikzlibrary{3d}

\usepackage{booktabs}       
\usepackage{array}          
\usepackage{multirow}       
\usepackage{makecell}       
\usepackage{longtable}      
\usepackage{threeparttable} 
\usepackage{tabularx}       
\usepackage{stackengine}    

\usepackage{xcolor}         
\definecolor{task1color}{RGB}{0,128,0}      
\definecolor{task2color}{RGB}{255,165,0}    
\definecolor{task3color}{RGB}{0,255,255}    
\colorlet{myred}{red!80!black}
\colorlet{myblue}{blue!80!black}
\colorlet{mygreen}{green!60!black}
\colorlet{mydarkblue}{blue!40!black}

\usepackage{colortbl}
\usepackage{array}
\usepackage{makecell}
\usepackage{caption}
\usepackage{subcaption}
\usepackage{adjustbox}

\definecolor{ourcolor}{RGB}{220,237,200}
\definecolor{bestcolor}{RGB}{255,215,0}

\usepackage{hyperref}       
\usepackage{url}            
\usepackage{cite}           
\usepackage[capitalize,noabbrev]{cleveref} 

\usepackage{balance}        
\usepackage{placeins}       
\usepackage{pdflscape}      
\usepackage{lscape}         
\usepackage{lipsum}         
\usepackage{textcomp}       

\usepackage{comment}        
\usepackage{enumitem}       
\usepackage{etoolbox}       
\usepackage{pifont}         

\graphicspath{{./Images/}}

\theoremstyle{plain}

\theoremstyle{definition}

\theoremstyle{remark}

\usepackage{algorithm}
\usepackage{algorithmic}

\title{No Forgetting Learning: Buffer-free Continual Learning Classification}

\author{
  Mohammad Ali Vahedifar and  Qi Zhang\thanks{This research was supported by the TOAST project, funded by the European Union’s Horizon Europe research and innovation program under the Marie Skłodowska-Curie Actions Doctoral Network (Grant Agreement No. 101073465), the Danish Council for Independent Research project eTouch (Grant No. 1127- 00339B), and NordForsk Nordic University Cooperation on Edge Intelligence (Grant No. 168043). } \\
  DIGIT and Department of Electrical and Computer Engineering, Aarhus University, Denmark \\
  \texttt{Authors' e-mails: \{av, qz\}@ece.au.dk.} \\
  \href{https://github.com/Ali-Vahedifar/No-Forgetting-Learning}{GitHub Code}
}

\begin{document}

\maketitle 
\begin{abstract}
Most Continual Learning (CL) methods maintain performance on earlier tasks by storing exemplars in a replay buffer, introducing memory overhead that scales with the number of tasks and raising privacy concerns in regulated domains. We propose \textbf{No Forgetting Learning (NFL)}, a buffer-free framework for class- and task-incremental learning that instead exploits the inherent redundancy of overparameterized networks.

NFL decomposes the network into a shared backbone and task-specific heads, then applies a \textbf{stepwise freezing} protocol: new capabilities are first isolated, shared representations are adapted under knowledge distillation, and all components are jointly refined with dual soft-target anchoring. \textbf{NFL+} augments this pipeline with an under-complete auto-encoder that preserves informative features from previous tasks and corrects the prediction bias caused by class imbalance. \textbf{NFL+LoRA} further extends the framework to pre-trained Vision Transformers by confining updates to a low-rank subspace with Fisher-weighted regularization, maintaining constant backbone memory cost regardless of the number of tasks. 

On CIFAR-100, Tiny-ImageNet, and ImageNet-1000 across up to 50 incremental tasks, NFL+ outperforms all buffer-free baselines and matches memory-based methods while requiring only \textbf{2.53\%} of their model size. We also propose a Plasticity--Stability score for more balanced trade-off evaluation.
\end{abstract}
\section{Introduction}
\label{sec:intro}
Neural Networks (NNs) trained sequentially on multiple tasks often suffer from Catastrophic Forgetting, in which new learning overwrites previously acquired knowledge~\citep{wang2023comprehensivesurveyforgettingdeep}. This failure mode reflects a fundamental tension between plasticity, the capacity to learn new tasks, and stability, the capacity to retain old ones. Continual learning (CL) seeks to resolve this by enabling models to absorb new knowledge without revisiting earlier data~\citep{vahedifar_2025_14631802}.

Most CL methods sidestep the problem by storing exemplars from previous tasks~\citep{rebuffi2017icarl, buzzega2020dark, Douillard_DyTox, zhou2023model}. While effective, this reliance on replay buffers introduces memory overhead that scales with the number of tasks, raises privacy concerns under regulations such as GDPR~\citep{voigt2017gdpr}, and violates the strict CL setting in which prior task data is inaccessible. Buffer-free alternatives exist, regularization-based methods penalize changes to important parameters~\citep{kirkpatrick2017overcoming, zenke2017continual}, and distillation-based methods transfer knowledge through soft targets~\citep{li2017learningforgetting}, but they consistently underperform their memory-based counterparts, particularly on large-scale benchmarks. The critical question is: \textit{How can we design a memory-efficient CL framework that operates within the fixed capacity of the backbone network without sacrificing performance?}

We answer affirmatively with \textbf{No Forgetting Learning (NFL)}, a memory-free CL framework built on a principled stepwise freezing strategy. NFL decomposes the network into a shared backbone and task-specific heads, then orchestrates which components are trained or frozen at each stage: isolating new task learning, adapting shared representations under distillation, and consolidating all components jointly. This staged approach prevents the gradient interference that causes forgetting in standard joint training. \textbf{NFL+} extends this pipeline with an under-complete auto-encoder that preserves the most informative features from previous tasks and corrects prediction bias toward new classes. To scale the framework to large pre-trained Vision Transformers (ViTs), we further propose \textbf{NFL+LoRA}, which replaces full-parameter updates with Low-Rank Adaptation (LoRA)~\citep{hu2022lora} and substitutes the auto-encoder with Fisher Information-based weight regularization in the full-dimensional update space~\citep{zheng2026revisiting}, maintaining constant backbone memory cost regardless of the number of tasks. Our contributions are as follows:

\textbf{(i)} We introduce NFL, a buffer-free CL framework that leverages stepwise freezing across systematically selected training configurations to mitigate catastrophic forgetting.

\textbf{(ii)} We propose NFL+, which augments NFL with an auto-encoder for feature preservation and a bias correction mechanism, and NFL+LoRA, which adapts the pipeline to pre-trained ViTs.

\textbf{(iii)} We propose a new evaluation metric that jointly captures the plasticity--stability trade-off, enabling more informative comparison across CL methods.

\textbf{Related work.} Due to space constraints, we defer our comprehensive review of prior work to Appendix~\ref{RW}. There, we provide an extended discussion of Continual Learning approaches.
\section{No Forgetting Learning}\label{Met}
We advocate that CL methods should avoid accessing data from previous tasks entirely. To this end, we propose a family of methods based on \textbf{stepwise freezing}: a principled training protocol that mitigates catastrophic forgetting by relying solely on the network's inherent parameters. We begin with the base method, NFL, then extend it with an auto-encoder in NFL+, and finally adapt the framework to Vision Transformers via NFL+LoRA~\citep{dosovitskiy2021image}.

\subsection{Design Philosophy}\label{design}
Standard compression techniques such as weight pruning~\citep{LTH} and neuron pruning~\citep{neuronghorbani} seek to reduce networks to 10--20\% of their original size. We investigate the converse question: \textit{can the inherent redundancy of overparameterized networks be repurposed to support continual learning?}

We hypothesize that by carefully orchestrating which parameters are updated at each training stage, the network's spare capacity can be used to stabilize learning rather than being pruned or overwritten. We decompose the network into three components: the shared backbone~$\theta_s$, the old task head~$\theta_t$, and the new task head~$\theta_{t+1}$. Each component is either trained~($T$) or frozen~($F$), yielding $2^3 = 8$ possible configurations. We analyze all eight to identify which are useful:
\begin{table}[h]
\centering
\small
\begin{tabular}{@{}ccccl@{}}
\toprule
$\#$ & $\theta_s$ & $\theta_t$ & $\theta_{t+1}$ & Outcome \\
\midrule
1 & $F$ & $F$ & $F$ & No learning occurs. \\
2 & $T$ & $F$ & $F$ & Backbone drifts without head adaptation; features misalign. \\
3 & $F$ & $T$ & $F$ & Old head re-fits on fixed features; no benefit for new task. \\
4 & $F$ & $T$ & $T$ & Heads update on a frozen backbone; insufficient plasticity. \\
5 & $T$ & $F$ & $T$ & Gradients from $\theta_{t+1}$ alter features needed by frozen $\theta_t$; directly causes forgetting. \\
\midrule
6 & $F$ & $F$ & $T$ & Safe initialization of the new head on existing features. \\
7 & $T$ & $T$ & $F$ & Controlled backbone adaptation anchored by old-task distillation. \\
8 & $T$ & $T$ & $T$ & Joint fine-tuning for global alignment; requires distillation safeguards. \\
\bottomrule
\end{tabular}
\end{table}

Configurations~1--5 are suboptimal. The remaining three (\#6, \#7, \#8) form the basis of our stepwise freezing pipeline. Crucially, their ordering matters: applying \#8 first would let the new task dominate the shared representation, while starting with \#6 ensures the new head is initialized without corrupting existing features. We therefore arrange them as follows:

\textbf{Step~1} trains the model on the current task $T_t$ using standard cross-entropy.

\textbf{Step~2} (\#6) freezes the backbone and old head, training only the new head $\theta_{t+1}$. This initializes the new classification capability without propagating noisy gradients into the backbone, addressing the failure mode identified in configuration~\#5.

\textbf{Step~3} (\#7) freezes the new head and updates the backbone and old head. With the new head frozen, the update is guided by distillation from the old task, ensuring features adapt only in directions compatible with previous decision boundaries.

\textbf{Step~4} (\#8) unfreezes all components for joint fine-tuning, stabilized by two sets of soft targets that anchor the representation to both the original and intermediate models.

\subsection{NFL: Base Method}\label{sec:nfl}

\begin{figure*}[t]
    \centering
    \includegraphics[width=\linewidth]{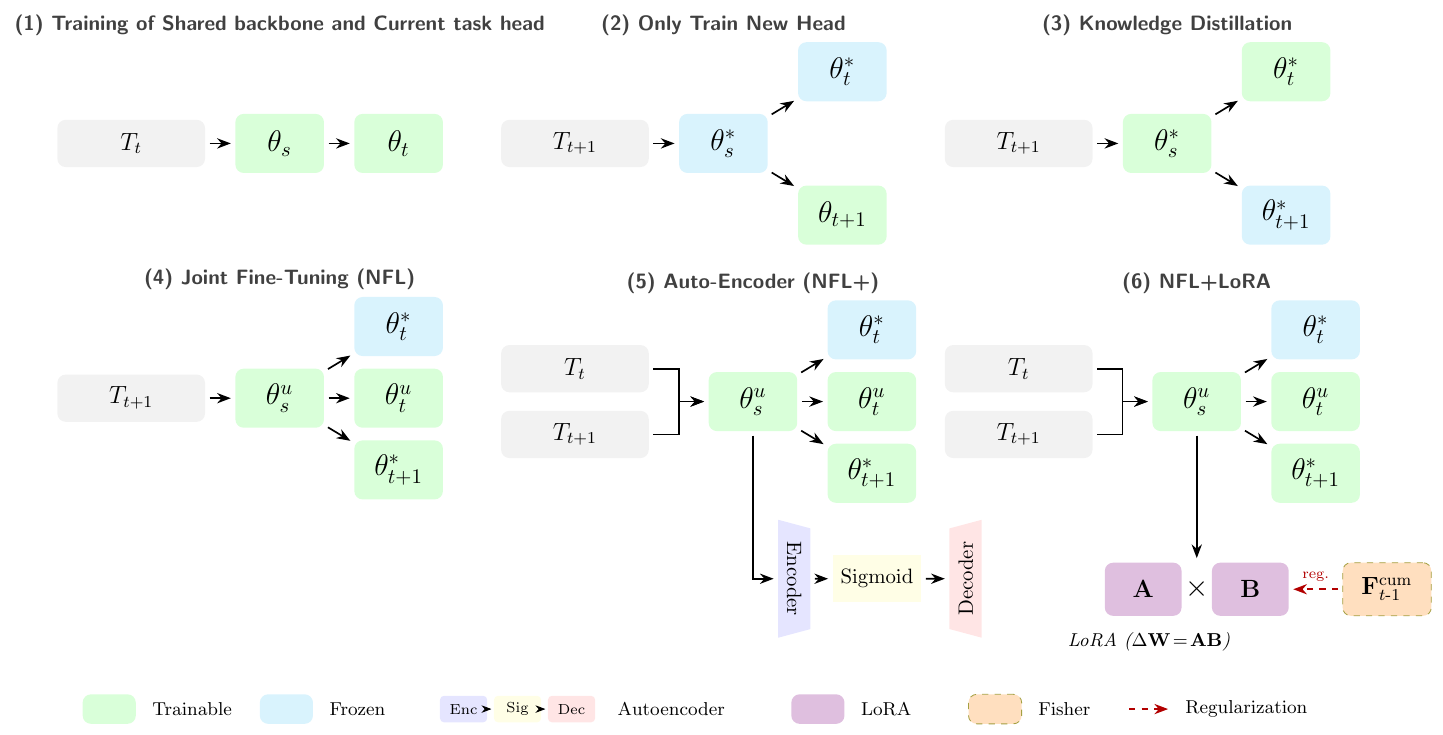}
    \caption{The stepwise freezing pipeline of NFL, NFL+, and NFL+LoRA. Each step's output serves as the next step's input. Steps~1--4 define the NFL base method; Step~5 adds the auto-encoder of NFL+; Step~6 adapts the framework to ViTs via NFL+LoRA.}
    \label{NFL concept}
\end{figure*}

We denote the model at step $i$ by $\text{NN}^i$ with outputs $O_k^i$, where subscript $k$ indexes the task head used to produce the output and superscript $i$ indexes the pipeline step. In \textbf{Step~1}, $\text{NN}^1(\theta_s, \theta_t)$ is trained on dataset $\mathcal{D}_t = (X_t, Y_t)$ for task $T_t$ with $c_t$ classes in class set $\mathcal{C}_t$, and $X_t$ as the data samples and $Y_{t}$ the ground-truth labels for task $T_{t}$ classes, using the standard cross-entropy loss:
\begin{equation}
    \mathcal{L}^1 = \mathbf{E}\Big[\mathcal{L}_{\text{CE}}\big(Y_t,\, O^1_t\big)\Big]. \label{loss 1}
\end{equation}
When a new task $T_{t+1}$ arrives with dataset $\mathcal{D}_{t+1} = (X_{t+1}, Y_{t+1})$, the goal is to learn the combined task $T_t \cup T_{t+1}$ over $\mathcal{C} = \mathcal{C}_t \cup \mathcal{C}_{t+1}$ without revisiting $\mathcal{D}_t$. To enable knowledge transfer, we compute soft targets by passing $X_{t+1}$ through the trained model:
\begin{equation}\label{eq: logit 1}
    H_t = \text{NN}^1\big(X_{t+1};\, \theta^*_s, \theta^*_t\big),
\end{equation}
where $H_t$ denotes the pre-softmax logits (i.e., the outputs before the softmax layer). These logits encode the old task's knowledge and will serve as distillation targets in subsequent steps.

In \textbf{Step~2}, we freeze $\theta^*_s$ and $\theta^*_t$ and train only the new head $\theta_{t+1}$:
\begin{equation}
    \mathcal{L}^2 = \mathbf{E}\Big[\mathcal{L}_{\text{CE}}\big(Y_{t+1},\, O^2_{t+1}\big)\Big]. \label{loss 2}
\end{equation}

In \textbf{Step~3}, we freeze $\theta^*_{t+1}$ obtained from Step~2 and update $\theta_s$ and $\theta_t$. The loss combines distillation against the stored logits $H_t$ with supervision on the new task:
\begin{equation}
    \mathcal{L}^3 = \mathbf{E}\Big[\mathcal{L}_{\text{KD}}\big(H_t,\, O^3_t\big) + \mathcal{L}_{\text{CE}}\big(Y_{t+1},\, O^3_{t+1}\big)\Big], \label{Loss 3}
\end{equation}
where the knowledge distillation loss is defined as:
\begin{equation}
    \mathcal{L}_{\text{KD}}(H, O) = -\sum_{i} \hat{h}_i \log \hat{o}_i, \qquad \hat{h}_i = \frac{\exp(H_i / p)}{\sum_j \exp(H_j / p)}, \qquad \hat{o}_i = \frac{\exp(O_i / p)}{\sum_j \exp(O_j / p)}.
    \label{eq:distillation}
\end{equation}
Here $p > 1$ is a temperature parameter that controls the smoothness of the softened distributions~\citep{hinton2015distillingknowledgeneuralnetwork}.

In \textbf{Step~4}, we unfreeze all components for joint fine-tuning. To prevent the new task from dominating the update, we distill against two sets of soft targets: the original logits $H_t$ from Step~1 and updated logits $H^\prime_t$ obtained by passing $X_{t+1}$ through the Step~3 backbone paired with the \emph{original} old-task head from Step~1:
\begin{equation}
    H^\prime_t = \text{NN}^4\big(X_{t+1};\, \theta^u_s, \theta^*_t\big), \label{EQ: H2}
\end{equation}
where $\theta^u_s$ denotes the backbone updated in Step~3 and $\theta^*_t$ denotes the original old-task head from Step~1 (not the updated head from Step~3). This combination captures how the old classifier responds to the adapted features, providing a complementary distillation anchor to $H_t$.

We denote the old-task head output of the current model at Step~4 as $O_t^4 = \theta_t(\theta_s(X_{t+1}))$. The final loss is:
\begin{equation}
    \mathcal{L}^4 = \mathbf{E}\Big[\alpha\,\mathcal{L}_{\text{KD}}\big(H_t,\, O_t^4\big) + (1 - \alpha)\,\mathcal{L}_{\text{KD}}\big(H^\prime_t,\, O_t^4\big) + \mathcal{L}_{\text{CE}}\big(Y_{t+1},\, O_{t+1}^4\big)\Big], \label{Loss 5}
\end{equation}
where $\alpha \in [0, 1]$ controls the balance between the two distillation anchors. Both KD terms use the \emph{same} student output $O_t^4$ from the old-task head, but compare it against different teacher signals: $H_t$ anchors the model to the original task boundary, while $H^\prime_t$ anchors it to the intermediate representation learned in Step~3. The full procedure is summarized in Algorithm~\ref{alg: NFL} in the Appendix.

\subsection{NFL+: Auto-Encoder Regularization}\label{sec:nflplus}
NFL+ extends the base method by introducing an auto-encoder that preserves informative features and corrects prediction bias caused by class imbalance. The pipeline follows the same four-step structure, with two additions: an auto-encoder trained at the \emph{end} of each task (while the current task's data is still available), and a bias correction applied in the final step (see Algorithm~\ref{alg: NFL+} and part~(5) of Fig.~\ref{NFL concept}).

\textbf{Feature preservation via auto-encoder.} After completing training on $T_t$ but \emph{before discarding} $\mathcal{D}_t$, we train an under-complete auto-encoder on the backbone's representations $P_t = \theta^*_s(X_t)$:
\begin{equation}
    R(P_t) = W_{\text{Dec}}\, \sigma\!\big(W_{\text{Enc}}\, P_t\big),
\end{equation}
where $W_{\text{Enc}}$ and $W_{\text{Dec}}$ are encoder and decoder weights, and $\sigma$ is an activation function. The auto-encoder learns a compressed summary of the feature manifold for $T_t$, capturing which directions in representation space are most important for the old task. The training objective is:
\begin{equation}
    \arg\min_{R}\; \mathbf{E}_{(X_t, Y_t)}\bigg[\Omega \big\| R(P_t) - P_t \big\|_2^2 + \mathcal{L}_{\text{CE}}\Big(\theta_t\big(R(P_t)\big),\, Y_t\Big)\bigg], \label{eq:ae_train}
\end{equation}
where $\Omega$ is a hyperparameter. The reconstruction term preserves fidelity, and the classification term ensures the compressed features remain discriminative. Crucially, this training occurs \emph{before} $T_{t+1}$ arrives: the auto-encoder is prepared at the end of each task and carried forward as a frozen module, fully consistent with the buffer-free constraint.

When $T_{t+1}$ arrives, $\mathcal{D}_t$ is discarded and the pipeline proceeds through Steps~1--4 as in NFL, with the auto-encoder providing the feature-space constraint and bias correction in Step~4.

\textbf{Bias correction.} In class-incremental learning, the model is exposed only to new-task data during adaptation, creating a systematic bias toward new classes. Because the auto-encoder's latent space captures the structure of old-task representations, it provides a natural basis for correcting this imbalance. We introduce a learned logit transformation $\Gamma$ that modulates the stored soft targets based on how the current input relates to the old feature manifold:
\begin{equation}
    \Gamma(P_{t+1}) = w_{\text{bias}}\, \sigma(W_{\text{Enc}}\, P_{t+1}) + b_{\text{bias}}, \qquad \widetilde{H}_t = \Gamma(P_{t+1}) \odot H_t, \label{adjust logits}
\end{equation}
where $P_{t+1} = \theta_s(X_{t+1})$ and $\odot$ is element-wise multiplication. The parameters $w_{\text{bias}}$ and $b_{\text{bias}}$ are trained on a held-out validation split of $\mathcal{D}_{t+1}$ (the 10\% cross-validation set described in Section~\ref{Experiements}), which contains only new-task data. All other model parameters remain frozen during this step. The training objective minimizes the distillation loss between the bias-corrected soft targets and the model's old-task output, subject to identity regularization:
\begin{equation}
    \arg\min_{w_{\text{bias}}, b_{\text{bias}}}\; \mathbf{E}_{X_{t+1}^{\text{val}}}\bigg[\mathcal{L}_{\text{KD}}\Big(\widetilde{H}_t,\, O_t\Big) + \mu\big\|\Gamma(P_{t+1}) - \mathbf{1}\big\|_2^2\bigg], \label{eq:bias_obj}
\end{equation}
where $\mu$ controls regularization strength. The KD term guides $\Gamma$ to rescale old-task logits so that the model's predictions better reflect the true class distribution, while the regularization term prevents the correction from deviating too far from the identity mapping. Note that this objective requires \emph{no old-task labels}: it operates entirely on the soft targets $H_t$ (stored as logits, not data) and new-task validation samples, preserving the buffer-free constraint.

In the final step, we additionally constrain the encoded representations to remain close to their values before adaptation, using the frozen encoder as a feature-space anchor:
\begin{equation}
    \mathbf{E}\Big[\big\|\sigma(W_{\text{Enc}}\, \theta_s(X_{t+1})) - \sigma(W_{\text{Enc}}\, \theta^*_s(X_{t+1}))\big\|_2^2\Big].
\end{equation}
We denote the old-task head output of the current model at Step~4 as $O_t^5 = \theta_t(\theta_s(X_{t+1}))$. The complete loss for the final step of NFL+ combines bias-corrected distillation, intermediate-model distillation, the feature-space constraint, and new-task supervision:
\begin{equation}
\hspace{-4mm}\begin{aligned}
    \mathcal{L}^5 = \mathbf{E}\bigg[
\underbrace{\eta\,\mathcal{L}_{\text{KD}}\Big(\widetilde{H}_t,\, O_t^5\Big)}_{\text{Bias-Corrected Stability}}
    + \underbrace{(1-\eta)\,\mathcal{L}_{\text{KD}}\Big(H^\prime_t,\, O_t^5\Big)}_{\text{Step 3 Model Stability}}
       + \underbrace{\Big\|\sigma\big(W_{\text{Enc}} \theta_s(X_{t+1})\big) - \sigma\big(W_{\text{Enc}} \theta^*_s(X_{t+1})\big) \Big\|_2^2}_{\text{Feature Space Constraint}}&\\
     +\underbrace{\mathcal{L}_{\text{CE}}\Big(Y_{t+1}, O_{t+1}^5\Big)}_{\text{New Task Plasticity}} \bigg]&.
\end{aligned}
\label{NFL+ loss 6}
\end{equation}
Here $\eta \in [0, 1]$ balances the two distillation anchors. As in NFL's Step~4 (Eq.~\ref{Loss 5}), both KD terms use the same student output $O_t^5$, compared against different teacher signals ($\widetilde{H}_t$ and $H^\prime_t$).

After completing adaptation to $T_{t+1}$, we retrain the auto-encoder on the current task's representations $P_{t+1} = \theta_s(X_{t+1})$ before discarding $\mathcal{D}_{t+1}$, preparing it for the next task transition (see Step~5 in Algorithm~\ref{alg: NFL+}).

\subsection{NFL+LoRA: Adaptation for Vision Transformers}\label{sec:nflpp}

The success of large-scale pre-trained Vision Transformers has shifted the CL paradigm toward adapting frozen models rather than training from scratch~\citep{zheng2026revisiting}. NFL+LoRA adapts the stepwise freezing philosophy to this setting by restricting all parameter updates to a low-rank subspace via LoRA~\citep{hu2022lora} (see Algorithm~\ref{alg: NFL+LoRA} and part~(6) of Fig.~\ref{NFL concept}).

The pre-trained weights $\mathbf{W}_0$ are used only for initialization. At each adapted layer, the weight update is parameterized as $\Delta\mathbf{W} = \mathbf{A}\mathbf{B}$, where $\mathbf{A} \in \mathbb{R}^{d_O \times r}$, $\mathbf{B} \in \mathbb{R}^{r \times d_I}$, and $r \ll \min(d_I, d_O)$. The effective weight after $t$ tasks is $\mathbf{W}_t = \mathbf{W}_{t-1} + \mathbf{A}^*\mathbf{B}^*$, where $\mathbf{W}_{t-1}$ denotes the \emph{accumulated} base weights incorporating all previously merged LoRA updates ($\mathbf{W}_0$ for the first task). Only $\mathbf{A}$ and $\mathbf{B}$ are trainable; $\mathbf{W}_{t-1}$ remains frozen throughout training on task $T_t$. When the backbone is marked as ``trainable'' in a given step (Steps~1, 3, 4), only the LoRA parameters are updated; when marked as ``frozen'' (Step~2), the LoRA parameters are also frozen.

The task lifecycle proceeds as follows. After completing training on task $T_t$, we: (i)~estimate the diagonal Fisher information matrix $\mathbf{F}_t$ over the learned LoRA parameters; (ii)~accumulate it as $\mathbf{F}_t^{\text{cum}} = \gamma\, \mathbf{F}_{t-1}^{\text{cum}} + \mathbf{F}_t$, where $\gamma$ is a decay factor; (iii)~merge the learned LoRA into the base weights as $\mathbf{W}_t = \mathbf{W}_{t-1} + \mathbf{A}^*\mathbf{B}^*$; and (iv)~initialize a fresh LoRA module $\mathbf{A}, \mathbf{B} \leftarrow \mathbf{0}$ for the next task. This merge-and-reinitialize scheme keeps the memory footprint constant regardless of the number of tasks. Since $\mathbf{W}_{t-1}$ is frozen during training, the gradient with respect to $\mathbf{W}$ and $\Delta\mathbf{W}$ are identical, so the Fisher matrix is computed in $\Delta\mathbf{W}$-space without additional overhead.

NFL+LoRA follows the same four-step structure as NFL. In Step~4, we denote the old-task head output as $O_t^4 = \theta_t(\theta_s(X_{t+1}))$ and apply the following loss, replacing the auto-encoder feature constraint of NFL+ with a Fisher-weighted regularization penalty:
\begin{equation}
    \mathcal{L}^{4} = \mathbf{E}\bigg[
    \alpha\,\mathcal{L}_{\text{KD}}\big(H_t,\, O_t^4\big)
    + (1 - \alpha)\,\mathcal{L}_{\text{KD}}\big(H^\prime_t,\, O_t^4\big)
    + \mathcal{L}_{\text{CE}}\big(Y_{t+1},\, O_{t+1}^4\big)
    + \frac{\lambda}{2}\,\text{vec}(\mathbf{A}\mathbf{B})^\top \mathbf{F}_{t}^{\text{cum}}\,\text{vec}(\mathbf{A}\mathbf{B})
    \bigg],
    \label{eq:nflpp_loss}
\end{equation}
where $\lambda$ controls regularization strength and $\text{vec}(\cdot)$ denotes vectorization. The first three terms mirror the dual-distillation structure of NFL's Step~4 (Eq.~\ref{Loss 5}), with both KD terms using the same student output $O_t^4$ compared against different teacher signals. The fourth term penalizes deviations in parameter directions that are important for previously learned tasks, using the accumulated Fisher $\mathbf{F}_{t}^{\text{cum}}$ (which incorporates information from all tasks through $T_t$). This serves the same stabilizing role as the auto-encoder constraint in NFL+ but through parameter-space rather than feature-space regularization.

\section{Experiment \& Discussion}\label{Experiements}

\textbf{Scenarios:} We evaluate performance under TIL and CIL settings (see section \ref{Evaluation Protocol} in Appendix).

\textbf{Evaluation Metrics:} To assess the ability of each method, we use Average Accuracy (ACC), Backward transfer (BWT), and Plasticity-Stability (PS) for the CIL and TIL scenarios (see section \ref{Evalution Metrices} in Appendix). 
In addition, we propose a new metric \textit{``Plasticity-Stability"}. The main idea is that effective CL requires a balance between plasticity, which allows the system to acquire new knowledge, and stability, which ensures that previously learned knowledge is retained. PS assesses how well CL methods scale with increasing numbers of tasks and classes. 

When CL methods trained for task $T_k$ on $D_k$, its accuracy on all tasks is measured using the corresponding test sets, leading to a matrix $A \in \mathbb{R}^{T \times T}$ containing the accuracies on all $T$ tasks, i.e., $A_{i,j}$ denotes the accuracy of the model on task $T_j$ after trained completely on task $T_i$.

\textbf{Plasticity-Stability}: This metric quantifies the trade-off between plasticity and stability:
\begin{align}
&\text{PS}_T = \frac{2 \cdot P \cdot S}{P + S},\quad
P = \frac{1}{T-1} \sum_{k=2}^{T} \frac{A_{k,k} - A_{k-1,k}}{1 - A_{k-1,k}},\quad
S = 1 - {\frac{1}{T-1} \sum_{k=1}^{T-1} (A_{k,k} - A_{T,k})}.
\end{align}
Here, $P$ measures learning efficiency by quantifying the extent to which the "unlearned" knowledge gap was closed during training. $S$ measures the retention rate of previously learned tasks. We assume $A_{k,k} \ge A_{T,k}$ (non-negative forgetting). If backward transfer increases accuracy ($A_{T,k} > A_{k,k}$), the forgetting term for that task is set to zero.

\textbf{Methods:} We compare our method's performance with \textit{Memory-free methods:} EWC~\citep{kirkpatrick2017overcoming}, SI~\citep{zenke2017continual}, LwF~\citep{li2017learningforgetting}, PEC~\citep{PEC}, SpaceNet~\citep{sokar2021spacenet}, NISPA~\citep{gurbuz2022nispa}, and DCNet~\citep{wang2025dcnet}. \textit{Memory-based methods:} iCaRL~\citep{rebuffi2017icarl}, DER++~\citep{buzzega2020dark}, DyTox~\citep{Douillard_DyTox}, and MEMO~\citep{zhou2023model}. \textit{LoRA-based methods (ViT-B/16):} CL-LoRA~\citep{he2025cllora} and EWC-LoRA~\citep{zheng2026revisiting}. We used Stochastic Gradient Descent (SGD) as Lower Bound (LB) and joint training as Upper Bound (UB).

\textbf{Datasets:} We conduct experiments on CIFAR-100~\citep{krizhevsky2009learning} (100 classes, 10/20 tasks, 10/5 classes/task), Tiny-ImageNet~\citep{TinyImageNet} (200 classes, 10/20 tasks, 20/10 classes/task), and ImageNet-1000~\citep{deng2009imagenet} (1,000 classes, 10/20/50 tasks, 100/50/20 classes/task) for the ResNet-18 backbone.

For the ViT-B/16 backbone, we additionally evaluate on ImageNet-R~\citep{hendrycks2021manyfaces} (200 classes, 10/20 tasks, 20/10 classes/task) and ImageNet-A~\citep{hendrycks2021natural} (200 classes, 10/20 tasks, 20/10 classes/task).

\textbf{Experiment Setup:} For experiments on the CIFAR-100, Tiny-ImageNet, and ImageNet-1000 datasets, we adopt the ResNet-18 architecture~\citep{He_2016_CVPR}. All network layers are initialized using He initialization~\citep{He_2015_ICCV}. To maintain consistency across CL methods, all models are trained with the Adam optimizer~\citep{adamkingma2017adammethodstochasticoptimization}. 

The Auto-Encoder is compact at 1.56 MB (3.5\%) compared to the main network's 44.68 MB. Table~\ref{tab:hyperparameters_complete} presents the full set of evaluated hyperparameter configurations for grid search, while Table~\ref{tab:best_hyperparameters} reports the best identified settings. All training is averaged over 10 runs, with 70\% of the data used for training, 20\% for testing, and 10\% reserved as a validation set. Training is performed for a maximum of 100 epochs on CIFAR-100 and Tiny-ImageNet, and 200 epochs on ImageNet-1000, with early stopping based on validation loss. The batch size is fixed to 64 for all experiments.

For memory-based methods in the CIL setting, we use 2,000 exemplars for CIFAR-100 and Tiny-ImageNet, and 20,000 exemplars for ImageNet-1000~\citep{zhou2024class}. In the TIL setting, we follow prior work~\citep{buzzega2020dark} and set the number of exemplars to 200 across all datasets. All experiments are conducted on a single NVIDIA A6000 GPU.

\textbf{Hyperparameters:} We adopt a tuning protocol aligned with the principles of Generalizable Two-phase Evaluation Protocol (GTEP)~\citep{cha2025hyperparameters} and the realistic hyperparameter optimization (HPO) recommendations of~\citet{lee2024hyperparameter} (See section \ref{Hyperparameter Search} in the Appendix). Hyperparameters are selected using only the first task's validation split (10\% of the first task's data) via grid search over the ranges specified in Table~\ref{tab:hyperparameters_complete}. Once selected, these hyperparameters are \emph{fixed for all subsequent tasks} and are never re-tuned on later data. 

This first-task HPO strategy avoids the two principal pitfalls identified in prior work: (i)~the conventional protocol of tuning and evaluating within the same scenario, which~\citet{cha2025hyperparameters} show leads to significant overestimation of CL capacity across more than 8{,}000 experiments; and (ii)~end-of-training HPO, which \citet{lee2024hyperparameter} demonstrates is unrealistic because it requires repeated passes over the entire task stream. We emphasize that this concern disproportionately affects memory-based methods, whose performance depends critically on buffer management hyperparameters (e.g., reservoir sampling rate, replay frequency, exemplar selection strategy). Tuning these parameters on the evaluation scenario implicitly leaks information about the distribution of future tasks into the buffer policy, precisely the information leakage that~\citet{cha2025hyperparameters} identify as the root cause of overestimation. NFL+ is \emph{structurally exempt} from this failure mode: as a memory-free method, it maintains no exemplar buffer and therefore has no buffer management hyperparameters to tune. Its hyperparameters govern the optimization dynamics rather than data selection and are determined entirely by the first task.
\begin{table*}[t]
\centering
\caption{Performance comparison on ImageNet-1000 across 10, 20, and 50 tasks (ResNet-18) for both CIL and TIL scenarios. Best results are highlighted for \textcolor{green!80}{buffer-free} and \textcolor{cyan!70}{memory-based}.}
\label{tab:imagenet_combined}
\vspace{0.5em}
\setlength{\tabcolsep}{2pt}
\renewcommand{\arraystretch}{1.05}
\resizebox{\textwidth}{!}{\begin{tabular}{@{}l ccc ccc ccc ccc ccc ccc@{}}
\toprule
& \multicolumn{9}{c}{\textbf{CIL}} & \multicolumn{9}{c}{\textbf{TIL}} \\
\cmidrule(lr){2-10} \cmidrule(lr){11-19}
& \multicolumn{3}{c}{\textbf{10 Tasks}} & \multicolumn{3}{c}{\textbf{20 Tasks}} & \multicolumn{3}{c}{\textbf{50 Tasks}}
& \multicolumn{3}{c}{\textbf{10 Tasks}} & \multicolumn{3}{c}{\textbf{20 Tasks}} & \multicolumn{3}{c}{\textbf{50 Tasks}} \\
\cmidrule(lr){2-4} \cmidrule(lr){5-7} \cmidrule(lr){8-10}
\cmidrule(lr){11-13} \cmidrule(lr){14-16} \cmidrule(lr){17-19}
\textbf{Method}
  & ACC$\uparrow$ & BWT$\uparrow$ & PS$\uparrow$
  & ACC$\uparrow$ & BWT$\uparrow$ & PS$\uparrow$
  & ACC$\uparrow$ & BWT$\uparrow$ & PS$\uparrow$
  & ACC$\uparrow$ & BWT$\uparrow$ & PS$\uparrow$
  & ACC$\uparrow$ & BWT$\uparrow$ & PS$\uparrow$
  & ACC$\uparrow$ & BWT$\uparrow$ & PS$\uparrow$ \\
\midrule
\multicolumn{19}{c}{\textit{\textbf{Memory-free methods}}} \\
NFL+ (Ours)
  &  \cellcolor{green!20}38.42 \scriptsize{±3.85} &  \cellcolor{green!20}$-$37.78 &  \cellcolor{green!20}0.67
  &  \cellcolor{green!20}31.50 \scriptsize{±3.62} &  \cellcolor{green!20}$-$41.20 &  \cellcolor{green!20}0.60
  &  \cellcolor{green!20}22.40 \scriptsize{±3.48} &  \cellcolor{green!20}$-$45.80 &  \cellcolor{green!20}0.51
  &  \cellcolor{green!20}51.36 \scriptsize{±4.52} & $-$5.19 &  \cellcolor{green!20}0.63
  &  \cellcolor{green!20}45.80 \scriptsize{±4.38} & $-$7.85 &  \cellcolor{green!20}0.56
  &  \cellcolor{green!20}37.20 \scriptsize{±4.15} & $-$12.40 &  \cellcolor{green!20}0.48 \\
DCNet
  & 37.80 \scriptsize{±3.72} & $-$38.96 & 0.66
  & 30.10 \scriptsize{±3.40} & $-$42.50 & 0.58
  & 20.80 \scriptsize{±3.25} & $-$47.30 & 0.48
  & 50.15 \scriptsize{±4.30} & $-$6.42 & 0.60
  & 44.30 \scriptsize{±4.12} & $-$9.50 & 0.53
  & 35.80 \scriptsize{±3.90} & $-$14.80 & 0.44 \\
WSN
  & --- & --- & ---
  & --- & --- & ---
  & --- & --- & ---
  & 48.73 \scriptsize{±1.85} &  \cellcolor{green!20}0.0 & 0.52
  & 43.20 \scriptsize{±1.72} &  \cellcolor{green!20}0.0 & 0.46
  & 35.40 \scriptsize{±1.58} &  \cellcolor{green!20}0.0 & 0.38 \\
NISPA
  & 29.40 \scriptsize{±4.15} & $-$43.08 & 0.61
  & 22.30 \scriptsize{±3.88} & $-$46.90 & 0.52
  & 14.50 \scriptsize{±3.60} & $-$51.40 & 0.41
  & 46.80 \scriptsize{±3.95} & $-$8.75 & 0.55
  & 40.60 \scriptsize{±3.72} & $-$12.30 & 0.47
  & 32.50 \scriptsize{±3.48} & $-$18.60 & 0.38 \\
NFL
  & 27.15 \scriptsize{±4.92} & $-$44.80 & 0.59
  & 21.40 \scriptsize{±4.55} & $-$48.50 & 0.50
  & 14.80 \scriptsize{±4.12} & $-$53.10 & 0.39
  & 40.18 \scriptsize{±5.32} & $-$23.62 & 0.49
  & 34.50 \scriptsize{±5.10} & $-$30.15 & 0.42
  & 26.80 \scriptsize{±4.78} & $-$38.70 & 0.34 \\
PEC
  & 14.83 \scriptsize{±4.25} & $-$50.20 & 0.53
  & 11.20 \scriptsize{±3.95} & $-$54.60 & 0.45
  & 7.40 \scriptsize{±3.50} & $-$59.20 & 0.35
  & --- & --- & ---
  & --- & --- & ---
  & --- & --- & --- \\
SpaceNet
  & 13.50 \scriptsize{±4.08} & $-$44.28 & 0.53
  & 10.40 \scriptsize{±3.76} & $-$48.50 & 0.44
  & 8.30 \scriptsize{±3.42} & $-$53.80 & 0.36
  & 37.20 \scriptsize{±4.15} & $-$18.40 & 0.40
  & 30.80 \scriptsize{±3.90} & $-$24.60 & 0.33
  & 23.10 \scriptsize{±3.65} & $-$32.40 & 0.25 \\
LwF
  & 11.24 \scriptsize{±4.38} & $-$47.50 & 0.51
  & 8.45 \scriptsize{±3.90} & $-$52.10 & 0.41
  & 5.30 \scriptsize{±3.15} & $-$57.40 & 0.30
  & 42.56 \scriptsize{±5.12} & $-$58.59 & 0.38
  & 35.20 \scriptsize{±4.85} & $-$65.30 & 0.31
  & 25.80 \scriptsize{±4.50} & $-$74.50 & 0.22 \\
SI
  & 9.68 \scriptsize{±5.45} & $-$45.51 & 0.50
  & 7.15 \scriptsize{±4.80} & $-$49.80 & 0.40
  & 4.50 \scriptsize{±3.85} & $-$55.20 & 0.28
  & 39.82 \scriptsize{±4.20} & $-$63.76 & 0.41
  & 33.40 \scriptsize{±4.05} & $-$70.20 & 0.34
  & 24.60 \scriptsize{±3.80} & $-$78.90 & 0.25 \\
EWC
  & 7.53 \scriptsize{±4.68} & $-$41.67 & 0.48
  & 5.60 \scriptsize{±4.20} & $-$46.20 & 0.38
  & 3.20 \scriptsize{±3.55} & $-$52.10 & 0.25
  & 36.47 \scriptsize{±4.02} & $-$54.13 & 0.37
  & 30.20 \scriptsize{±3.88} & $-$61.50 & 0.30
  & 22.10 \scriptsize{±3.62} & $-$70.80 & 0.22 \\
\midrule
\multicolumn{19}{c}{\textit{\textbf{Memory-based methods}}} \\
DyTox
  & \cellcolor{cyan!20}40.15 \scriptsize{±4.30} & $-$35.20 & 0.58
  & \cellcolor{cyan!20}33.20 \scriptsize{±4.12} & $-$38.80 & 0.50
  & \cellcolor{cyan!20}24.50 \scriptsize{±3.90} & $-$43.50 & 0.42
  & \cellcolor{cyan!20}59.40 \scriptsize{±4.65} & $-$6.15 & 0.58
  & \cellcolor{cyan!20}52.10 \scriptsize{±4.48} & $-$8.90 & 0.50
  & \cellcolor{cyan!20}42.80 \scriptsize{±4.20} & $-$13.70 & 0.42 \\
MEMO
  & 38.90 \scriptsize{±4.52} & $-$37.50 & \cellcolor{cyan!20}0.62
  & 31.80 \scriptsize{±4.28} & $-$41.20 & \cellcolor{cyan!20}0.55
  & 23.10 \scriptsize{±4.05} & $-$46.10 & \cellcolor{cyan!20}0.46
  & 58.25 \scriptsize{±4.80} & \cellcolor{cyan!20}$-$5.40 & \cellcolor{cyan!20}0.62
  & 50.80 \scriptsize{±4.62} & \cellcolor{cyan!20}$-$7.85 & \cellcolor{cyan!20}0.55
  & 41.50 \scriptsize{±4.35} & \cellcolor{cyan!20}$-$12.20 & \cellcolor{cyan!20}0.46 \\
DER++
  & 14.20 \scriptsize{±5.15} & \cellcolor{cyan!20}$-$32.80 & 0.36
  & 10.35 \scriptsize{±4.80} & \cellcolor{cyan!20}$-$36.50 & 0.30
  & 6.10 \scriptsize{±4.25} & \cellcolor{cyan!20}$-$41.80 & 0.23
  & 55.80 \scriptsize{±4.72} & $-$6.30 & 0.36
  & 48.20 \scriptsize{±4.50} & $-$9.15 & 0.30
  & 39.40 \scriptsize{±4.18} & $-$14.50 & 0.23 \\
iCaRL
  & 12.45 \scriptsize{±5.70} & $-$56.40 & 0.37
  & 8.80 \scriptsize{±5.25} & $-$60.20 & 0.31
  & 5.30 \scriptsize{±4.65} & $-$65.50 & 0.24
  & 55.10 \scriptsize{±4.35} & $-$9.80 & 0.37
  & 47.50 \scriptsize{±4.15} & $-$13.60 & 0.31
  & 38.20 \scriptsize{±3.90} & $-$19.40 & 0.24 \\
\bottomrule
\end{tabular}}
\end{table*}
\subsection{ImageNet-1000 results}

As shown in Table~\ref{tab:imagenet_combined}, the CIL setting, NFL+, achieves the highest accuracy among all memory-free methods, reaching 38.42\% on 10 tasks and maintaining a lead of 0.62, 1.40, and 1.60 percentage points over the nearest competitor, DCNet, at 10, 20, and 50 tasks, respectively. Notably, the advantage of NFL+ widens as the number of tasks increases, indicating that the multi-step freezing strategy scales more gracefully than dynamic channel expansion alone. NFL+ also achieves the best BWT among memory-free methods ($-37.78$ on 10 tasks), confirming that the freezing schedule effectively mitigates catastrophic forgetting. Compared to memory-based methods, NFL+ narrows the gap to DyTox, the overall best performer, to within 1.73 points across 10 tasks, despite using no exemplar buffer.

In the TIL setting, NFL+ again leads all memory-free methods across all configurations, achieving 51.36\%, 45.80\%, and 37.20\% ACC at 10, 20, and 50 tasks, respectively. An instructive comparison is with WSN, which achieves zero forgetting by construction (BWT\,=\,0.0) through hard weight masking but sacrifices plasticity, resulting in lower ACC and PS than NFL+ at every task count. This highlights that NFL+'s freezing strategy achieves a more favorable stability--plasticity tradeoff.

A cross-scenario comparison reveals that the relative ranking of methods shifts considerably between CIL and TIL. Methods like EWC and SI, which perform poorly in CIL (7.53\% and 9.68\% at 10 tasks), recover substantially in TIL (36.47\% and 39.82\%), reflecting the known difficulty gap between the two evaluation protocols. NFL+'s advantage is most pronounced in the harder CIL setting, where the absence of task identity at inference makes inter-task discrimination critical, precisely the scenario where the multi-step freezing and knowledge distillation pipeline provides the greatest benefit.
\begin{figure*}[t]
    \centering
    \includegraphics[width=\textwidth]{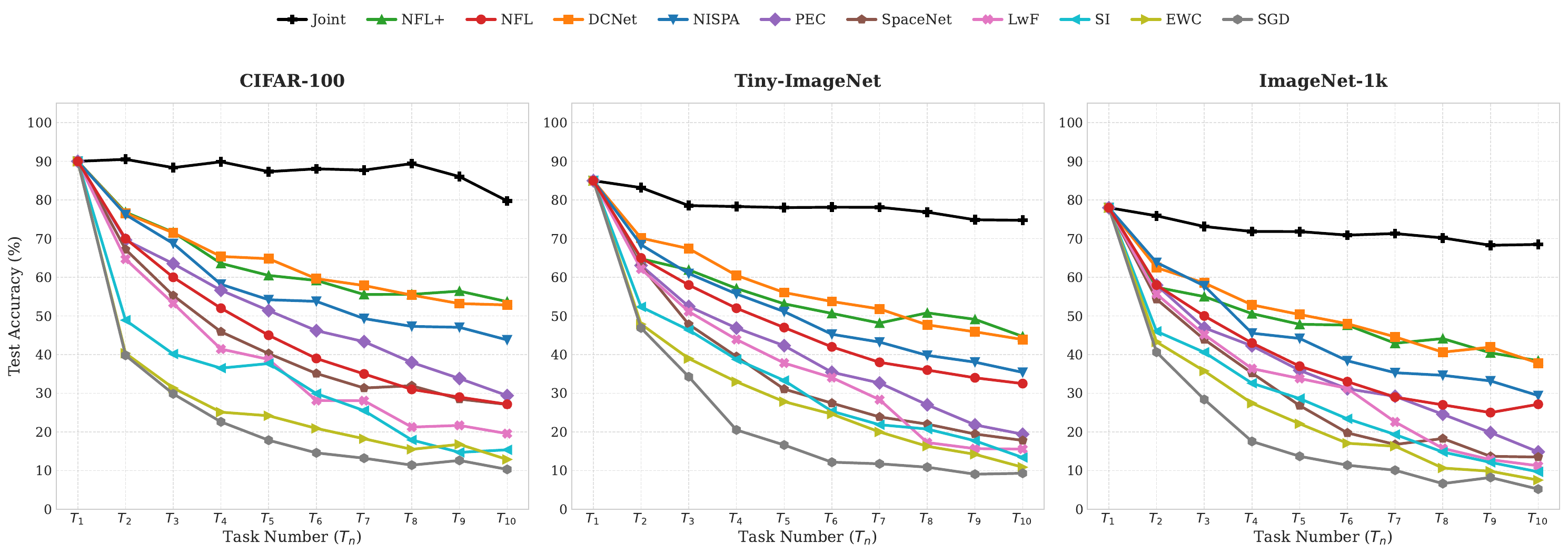}
    \caption{ACC evaluation for the CIL across 10 tasks. Each point represents the average accuracy evaluated after learning a given task, averaged over all tasks learned up to that point. For example, the value for task 5 corresponds to the average accuracy of the model on the test sets for tasks 1 through 5 after training on task 5 (see Fig.~\ref{fig: 10task_NFL+_matrix}). Solid lines: memory-based; dashed lines: buffer-free.}
    \label{fig:CIL10task}
\end{figure*}

\textbf{Comparison based on the CIL Scenario.} For accuracy evaluation, we report the average accuracy over all tasks learned up to the current stage, as shown in Fig.~\ref{fig:CIL10task}. All methods start from the same level, as no CL has occurred at the initial stage. As new tasks are introduced, performance degrades across methods, with dynamic-architecture approaches exhibiting greater robustness due to model expansion. Despite this advantage, NFL+ achieves competitive and, in several cases, superior performance in the CIL setting, particularly when compared to iCaRL and DER++. 

Although MEMO and DyTox attain higher accuracy, they rely on substantially larger model capacities or memory budgets. These results indicate that even with expanded architectures and privileged access to past data, such methods do not uniformly dominate more parameter-efficient approaches. In contrast, NFL+ demonstrates effective use of parameters and maintains strong performance despite operating in a more restrictive, arguably less favorable evaluation setting. Finally, Fig.~\ref{fig:CIL10task} reveals a persistent gap between all CL methods and the upper bound, highlighting the ongoing challenge of achieving robust long-term knowledge retention and transfer in CL.
\subsection{Architectural generalization to ViT-B/16}
As shown in Table~\ref{tab:CIL_vit}, NFL+LoRA achieves the highest ACC and lowest forgetting across all datasets and task configurations. On CIFAR-100 (10 tasks), NFL+LoRA reaches 90.80\% with a BWT of $-$1.85, outperforming the next best method EWC-LoRA by 2.89 points in ACC and an order of magnitude in forgetting. The advantage becomes more pronounced on the distribution-shifted benchmarks: on ImageNet-A at 20 tasks, NFL+LoRA achieves 59.10\% compared to 55.20\% for EWC-LoRA and 52.80\% for CL-LoRA, a gap of 3.90 and 6.30 points, respectively.

Two observations merit discussion. First, the BWT of NFL+LoRA demonstrates that the multi-step freezing strategy transfers effectively from full fine-tuning to parameter-efficient adaptation. The Fisher-regularized LoRA provides structural protection against forgetting that neither naive continual LoRA (CL-LoRA) nor EWC-based regularization can match. Second, the relative ranking between CL-LoRA and EWC-LoRA is dataset-dependent: CL-LoRA leads on ImageNet-R (78.72 vs.\ 72.86 at 10 tasks) while EWC-LoRA leads on ImageNet-A (59.89 vs.\ 57.62). This instability suggests that the effectiveness of regularization-based approaches is sensitive to the distributional characteristics of each benchmark. In contrast, NFL+LoRA dominates uniformly, indicating that the freezing-based design is more robust to distribution shift than penalty-based alternatives.
\begin{table*}[t]
\centering
\caption{Comparison of LoRA-based CIL methods using a ViT-B/16 backbone pretrained on ImageNet-21K. Best results are highlighted in \textcolor{violet!70}{violet}.}
\label{tab:CIL_vit}
\vspace{0.5em}
\small
\setlength{\tabcolsep}{3.5pt}
\renewcommand{\arraystretch}{1.05}
\resizebox{\textwidth}{!}{\begin{tabular}{@{}l ccc ccc ccc ccc ccc@{}}
\toprule
& \multicolumn{3}{c}{\textbf{CIFAR-100}} & \multicolumn{6}{c}{\textbf{ImageNet-R}} & \multicolumn{6}{c}{\textbf{ImageNet-A}} \\
\cmidrule(lr){2-4} \cmidrule(lr){5-10} \cmidrule(lr){11-16}
& \multicolumn{3}{c}{10 Tasks} & \multicolumn{3}{c}{10 Tasks} & \multicolumn{3}{c}{20 Tasks} & \multicolumn{3}{c}{10 Tasks} & \multicolumn{3}{c}{20 Tasks} \\
\cmidrule(lr){2-4} \cmidrule(lr){5-7} \cmidrule(lr){8-10} \cmidrule(lr){11-13} \cmidrule(lr){14-16}
\textbf{Method} & ACC$\uparrow$ & BWT$\uparrow$ & PS$\uparrow$ & ACC$\uparrow$ & BWT$\uparrow$ & PS$\uparrow$ & ACC$\uparrow$ & BWT$\uparrow$ & PS$\uparrow$ & ACC$\uparrow$ & BWT$\uparrow$ & PS$\uparrow$ & ACC$\uparrow$ & BWT$\uparrow$ & PS$\uparrow$ \\
\midrule
\multicolumn{16}{c}{\textit{\textbf{Continual Learning Only}}} \\
CL-LoRA
  & 87.65\scriptsize{±0.53} & $-$5.80 & 0.70
  & 78.72\scriptsize{±0.44} & $-$7.20 & 0.67
  & 74.30\scriptsize{±0.62} & $-$11.40 & 0.61
  & 57.62\scriptsize{±0.89} & $-$10.40 & 0.58
  & 52.80\scriptsize{±1.05} & $-$18.50 & 0.50 \\
EWC-LoRA
  & 87.91\scriptsize{±0.57} & $-$4.50 & 0.73
  & 72.86\scriptsize{±0.79} & $-$8.60 & 0.65
  & 68.15\scriptsize{±0.88} & $-$13.20 & 0.57
  & 59.89\scriptsize{±0.26} & $-$8.50 & 0.62
  & 55.20\scriptsize{±0.42} & $-$14.80 & 0.54 \\
  \midrule
NFL+LoRA (Ours)
  & \cellcolor{violet!10}90.80\scriptsize{±0.41} & \cellcolor{violet!10}$-$1.85 & \cellcolor{violet!10}0.72
  & \cellcolor{violet!10}79.70\scriptsize{±0.55} & \cellcolor{violet!10}$-$2.40 & \cellcolor{violet!10}0.66
  & \cellcolor{violet!10}76.15\scriptsize{±0.68} & \cellcolor{violet!10}$-$4.50 & \cellcolor{violet!10}0.62
  & \cellcolor{violet!10}63.40\scriptsize{±0.48} & \cellcolor{violet!10}$-$3.80 & \cellcolor{violet!10}0.61
  & \cellcolor{violet!10}59.10\scriptsize{±0.58} & \cellcolor{violet!10}$-$6.90 & \cellcolor{violet!10}0.56 \\
\bottomrule
\end{tabular}}
\end{table*}

\textbf{Appendix.} Additional results are in the Appendix, including a comparison based on the CIL in section~\ref{section CIL CifarTiny} and TIL~\ref{Section TIL cifarTiny} scenarios for CIFAR-100 and Tiny-ImageNet, analysis of the NFL+ Accuracy Matrix in section~\ref{Analysis of the NFL+}, task Granularity Analysis for the CIL in section~\ref{Task Granularity Analysis for CIL}, memory-wise comparison in section~\ref{section Memory-wise comparison}, computation cost in section~\ref{section Computational Cost Comparison}, parameter usage efficiency in section~\ref{Parameter Usage Efficiency} and ablation study of each step's contribution for the NFL and the NFL+ in section~\ref{Ablation Study of the NFL and the NFL+}.

\section{Conclusion}\label{Conclusion}

This work demonstrated that the inherent redundancy of overparameterized networks can be systematically exploited for CL without any replay buffer. The stepwise freezing protocol at the core of NFL, NFL+, and NFL+LoRA provides a principled alternative to the dominant paradigm of exemplar storage: by controlling which components are trained at each stage, gradient interference is mitigated at its source rather than compensated after the fact. The empirical results confirm that this approach closes much of the gap with memory-based methods while requiring only a fraction of their storage overhead, and the proposed Plasticity--Stability score offers a more informative lens for comparing methods that navigate this trade-off differently.

\textbf{Limitations.}
The reliance on knowledge distillation introduces a known failure mode: when successive tasks are drawn from substantially different distributions, the soft targets computed from old-task logits become increasingly unreliable, degrading the distillation signal. This is an inherent limitation of any distillation-based CL method, but it is particularly relevant in the buffer-free setting where no stored exemplars are available to recalibrate the targets. In NFL+, the auto-encoder adds a per-task training overhead and introduces a bottleneck dimension that must be chosen appropriately; while we found this choice to be robust across our benchmarks, the optimal compression ratio may vary for tasks with very different feature complexities. Finally, the multi-step pipeline increases the per-task training cost relative to single-pass methods; the computational analysis in Section~\ref{section Computational Cost Comparison} shows this remains practical, but it is a trade-off that users should weigh against their deployment constraints.

\textbf{Things We Tried That Did Not Work.} A comprehensive discussion of the suboptimal strategies and limitations encountered during this study is presented in Section~\ref{Things We Tried That Did Not Work}.

\textbf{Future work.}
A natural extension is adapting the stepwise freezing framework to large language models for text-based CL, where the tension between retaining linguistic knowledge and acquiring new capabilities is acute, and replay buffers are prohibitively expensive at scale. The LoRA-based variant already provides a starting point: its merge-and-reinitialize scheme and Fisher-weighted regularization are architecture-agnostic and directly applicable to transformer-based language models. Investigating how the distillation signal can be made more robust under distribution shift, for instance through adaptive temperature scaling or task-aware target selection, is another promising direction for strengthening the framework's applicability across diverse CL scenarios.

\bibliographystyle{plainnat}
\bibliography{paper}

@String(CVPR= {IEEE Conf. Comput. Vis. Pattern Recog.})

@String(ICCV= {Int. Conf. Comput. Vis.})

@String(NeurIPS= {Adv. Neural Inform. Process. Syst.})

@String(ICLR = {Int. Conf. Learn. Represent.})

@String(IJCAI = {IJCAI})

@String(TMLR = {Transactions on Machine Learning Research})

@String(CVPR  = {CVPR})

@String(ICCV  = {ICCV})

@String(NeurIPS  = {NeurIPS})

@String(ICLR  = {ICLR})

@inproceedings{zenke2017continual,
  title={{Continual learning through synaptic intelligence}},
  author={Zenke, Friedemann and Poole, Ben and Ganguli, Surya},
  booktitle={ICML},
  year={2017}
}

@inproceedings{PEC,
    title={Prediction Error-based Classification for Class-Incremental Learning},
    author={Michal Zajac and Tinne Tuytelaars and Gido M van de Ven},
    booktitle={ICLR},
    year={2024}
}

@article{voigt2017gdpr,
  title={The EU General Data Protection Regulation (GDPR)},
  author={Voigt, Paul and Von dem Bussche, Axel},
  journal={A Practical Guide, 1st Ed., Cham: Springer International Publishing},
  year={2017}
}

@InProceedings{haeyong,
  title = 	 {Forget-free Continual Learning with Winning Subnetworks},
  author = {Kang, Haeyong and others},
  booktitle = 	 {ICML},
  year = 	 {2022}
}

@misc{hinton2015distillingknowledgeneuralnetwork,
      title={Distilling the Knowledge in a Neural Network}, 
      author={Geoffrey Hinton and Oriol Vinyals and Jeff Dean},
      year={2015},
       note={https://arxiv.org/abs/1503.02531}, 
}

@article{kirkpatrick2017overcoming,
  title={Overcoming Catastrophic Forgetting in Neural Networks},
  author={Kirkpatrick, James and Pascanu, Razvan and Rabinowitz, Neil and Veness, Joel and Desjardins, Guillaume and Rusu, Andrei A. and Milan, Kieran and Quan, John and Ramalho, Tiago and Grabska-Barwinska, Agnieszka and Hassabis, Demis and Clopath, Claudia and Kumaran, Dharshan and Hadsell, Raia},
  journal={Proceedings of the National Academy of Sciences},

  year={2017}
}

@article{li2017learningforgetting,
  author={Li, Zhizhong and Hoiem, Derek},
  journal={TPAMI}, 
  title={Learning without Forgetting}, 
  year={2018}}

@InProceedings{He_2015_ICCV,
author = {He, Kaiming and Zhang, Xiangyu and Ren, Shaoqing and Sun, Jian},
title = {Delving Deep into Rectifiers: Surpassing Human-Level Performance on ImageNet Classification},
booktitle = {ICCV},
year = {2015},
}

@inproceedings{deng2009imagenet,
  title={ImageNet: A Large-Scale Hierarchical Image Database},
  author={Deng, Jia and Dong, Wei and Socher, Richard and Li, Li-Jia and Li, Kai and Fei-Fei, Li},
  booktitle={CVPR},
  year={2009},
  pages={248--255},
  doi={10.1109/CVPR.2009.5206848}
}

@article{wickramasinghe2024continual,
  title={Continual Learning: A Review of Techniques, Challenges and Future Directions},
  author={Wickramasinghe, Buddhi and Saha, Gobinda and Roy, Kaushik},
  journal={TAI},
  year={2024}
}

@article{wang2023comprehensivesurveyforgettingdeep,
      title={A Comprehensive Survey of Forgetting in Deep Learning Beyond Continual Learning}, 
      author={Zhenyi Wang and Enneng Yang and Li Shen and Heng Huang},
      year={2025},
      journal={TPAMI},  
}

@inproceedings{rebuffi2017icarl,
  title={iCaRL: Incremental classifier and representation learning},
  author={Rebuffi, Sylvestre-Alvise and Kolesnikov, Alexander and Sperl, Georg and Lampert, Christoph H.},
  booktitle={CVPR},
  year={2017}
}

@inproceedings{adamkingma2017adammethodstochasticoptimization,
  author = {Kingma, Diederik P. and Ba, Jimmy},
  booktitle = {ICLR},
  title = {Adam: A Method for Stochastic Optimization.},
  year = {2015}
}

@InProceedings{LTH,
 title = 	 {Linear Mode Connectivity and the Lottery Ticket Hypothesis},
 author =  {Frankle, Jonathan and others},
 booktitle = 	 {ICML},
 year = 	 {2020}
}

@misc{TinyImageNet,
  title={Tiny ImageNet Visual Recognition Challenge},
  author={Ya Le and Xuan S. Yang},
  year={2015}
}

@inproceedings{buzzega2020dark,
  title={Dark experience for general continual learning: a strong, simple baseline},
  author={Buzzega, Pietro and Boschini, Matteo and Porrello, Angelo and Abati, Davide and Calderara, Simone},
  booktitle={NeurIPS},
  year={2020}
}

@techreport{krizhevsky2009learning,
  title={Learning multiple layers of features from tiny images},
  author={Krizhevsky, Alex},
  year={2009},
  institution={University of Toronto},
  type={Technical report}
}

@InProceedings{He_2016_CVPR,
author = {He, Kaiming and Zhang, Xiangyu and Ren, Shaoqing and Sun, Jian},
title = {Deep Residual Learning for Image Recognition},
booktitle = {CVPR},
year = {2016}
}

@inproceedings{Lopez-PazNeurIPS2017_f8752278,
 author = {Lopez-Paz, David and Ranzato, Marc\textquotesingle Aurelio},
 booktitle = {NeurIPS},
 editor = {I. Guyon and U. Von Luxburg and S. Bengio and H. Wallach and R. Fergus and S. Vishwanathan and R. Garnett},
 title = {Gradient Episodic Memory for Continual Learning},
 volume = {30},
 year = {2017}
}

@InProceedings{diazrodriguez2018dontforgetforgettingnew,
      title={Don't forget, there is more than forgetting: new metrics for Continual Learning}, 
      author={Natalia Díaz-Rodríguez and Vincenzo Lomonaco and David Filliat and Davide Maltoni},
      year={2018},
      booktitle={NeurIPS workshop on Continual Learning} 
}

@article{zhou2024class,
    author = {Zhou, Da-Wei and Wang, Qi-Wei and Qi, Zhi-Hong and Ye, Han-Jia and Zhan, De-Chuan and Liu, Ziwei},
    title = {Class-Incremental Learning: A Survey},
    journal={TPAMI},
    year = {2024}
 }

@misc{vahedifar_2025_14631802,
  author       = {Vahedifar, Mohammad Ali and
                  Zhang, Qi and
                  Iosifidis, Alexandros},
  title        = {Towards Lifelong Deep Learning: A Review of
                   Continual Learning and Unlearning Methods
                  },
  year         = 2025,
  publisher    = {Zenodo},
  note          = {https://doi.org/10.5281/zenodo.14631802},
}

@inproceedings{neuronghorbani,
 author = {Ghorbani, Amirata and Zou, James Y},
 booktitle = {NeurIPS},
 title = {Neuron Shapley: Discovering the Responsible Neurons},
 year = {2020}
}

@InProceedings{Douillard_DyTox,
    author    = {Douillard, Arthur and Ram\'e, Alexandre and Couairon, Guillaume and Cord, Matthieu},
    title     = {{DyTox: Transformers for Continual Learning With DYnamic TOken eXpansion}},
    booktitle = {CVPR},
    year      = {2022}
}

@inproceedings{zhou2024continuallearningpretrainedmodels,
      title={{Continual Learning with Pre-Trained Models: A Survey}}, 
      author={Da-Wei Zhou and Hai-Long Sun and Jingyi Ning and Han-Jia Ye and De-Chuan Zhan},
      year={2024},
      booktitle={IJCAI}
}

@article{sokar2021spacenet,
  title     = {{SpaceNet}: Make Free Space for Continual Learning},
  author    = {Sokar, Ghada and Mocanu, Decebal Constantin and Pechenizkiy, Mykola},
  journal   = {Neurocomputing},
   year      = {2021}
}

@inproceedings{gurbuz2022nispa,
  title     = {{NISPA}: Neuro-Inspired Stability-Plasticity Adaptation for Continual Learning in Sparse Networks},
  author    = {Gurbuz, Mustafa B and Dovrolis, Constantine},
  booktitle = {ICML},
  pages     = {8064--8094},
  year      = {2022},
  organization = {PMLR}
}

@inproceedings{wang2025dcnet,
  title     = {On the Discrimination and Consistency for Exemplar-Free Class Incremental Learning},
  author    = {Wang, Tianqi and Guo, Jingcai and Li, Depeng and Chen, Zhi},
  booktitle = {IJCAI},
  pages     = {6424--6432},
  year      = {2025}
}

@inproceedings{L2p,
  title     = {Learning to Prompt for Continual Learning},
  author    = {Wang, Zifeng and Zhang, Zizhao and Lee, Chen-Yu and Zhang, Han and Sun, Ruoxi and Ren, Xiaoqi and Su, Guolong and Perot, Vincent and Dy, Jennifer and Pfister, Tomas},
  booktitle = {CVPR},
  pages     = {139--149},
  year      = {2022}
}

@inproceedings{hendrycks2021manyfaces,
  title     = {The Many Faces of Robustness: A Critical Analysis of Out-of-Distribution Generalization},
  author    = {Hendrycks, Dan and Basart, Steven and Mu, Norman and Kadavath, Saurav and Wang, Frank and Dorundo, Evan and Desai, Rahul and Zhu, Tyler and Parajuli, Samyak and Guo, Mike and others},
  booktitle = {ICCV},
  pages     = {8340--8349},
  year      = {2021}
}

@article{cha2025hyperparameters,
  title     = {Hyperparameters in Continual Learning: A Reality Check},
  author    = {Cha, Sungmin and Cho, Kyunghyun},
  journal   = {TMLR},
  year      = {2025}
}

@article{lee2024hyperparameter,
  title     = {Hyperparameter Selection in Continual Learning},
  author    = {Lee, Thomas L. and Hellan, Sigrid Passano and Ericsson, Linus and Crowley, Elliot J. and Storkey, Amos},
  journal   = {arXiv preprint arXiv:2404.06466},
  year      = {2024},
  url       = {https://arxiv.org/abs/2404.06466}
}

@inproceedings{hendrycks2021natural,
  title     = {Natural Adversarial Examples},
  author    = {Hendrycks, Dan and Zhao, Kevin and Basart, Steven and Steinhardt, Jacob and Song, Dawn},
  booktitle = {CVPR},
  year      = {2021}
}

@inproceedings{Coda,
  title     = {{CODA-Prompt}: Continual Decomposed Attention-Based Prompting for Rehearsal-Free Continual Learning},
  author    = {Smith, James Seale and Karlinsky, Leonid and Gutta, Vyshnavi and Cascante-Bonilla, Paola and Kim, Donghyun and Arbelle, Assaf and Panda, Rameswar and Feris, Rogerio and Kira, Zsolt},
  booktitle = {CVPR},
  year      = {2023}
}

@inproceedings{he2025cllora,
  title     = {{CL-LoRA}: Continual Low-Rank Adaptation for Rehearsal-Free Class-Incremental Learning},
  author    = {He, Jiangpeng and Duan, Zhihao and Zhu, Fengqing},
  booktitle = {CVPR},
  year      = {2025}
}

@inproceedings{dosovitskiy2021image,
  title     = {An Image is Worth 16x16 Words: Transformers for Image Recognition at Scale},
  author    = {Dosovitskiy, Alexey and Beyer, Lucas and Kolesnikov, Alexander and Weissenborn, Dirk and Zhai, Xiaohua and Unterthiner, Thomas and Dehghani, Mostafa and Minderer, Matthias and Heigold, Georg and Gelly, Sylvain and Uszkoreit, Jakob and Houlsby, Neil},
  booktitle = {ICLR},
  year      = {2021}
}

@inproceedings{hu2022lora,
  title     = {LoRA: Low-Rank Adaptation of Large Language Models},
  author    = {Hu, Edward J. and Shen, Yelong and Wallis, Phillip and Allen-Zhu, Zeyuan and Li, Yuanzhi and Wang, Shean and Wang, Lu and Chen, Weizhu},
  booktitle = {ICLR},
  year      = {2022}
}

@inproceedings{zheng2026revisiting,
  title   = {Revisiting Weight Regularization for Low-Rank Continual Learning},
  author  = {Zheng, Yaoyue and Zhang, Yin and van de Weijer, Joost and van de Ven, Gido M. and Du, Shaoyi and Zhang, Xuetao and Tian, Zhiqiang},
  year    = {2026},
  booktitle = {ICLR}
}

@article{qu2024recentadvancescontinuallearning,
      title={{Recent Advances of Continual Learning in Computer Vision: An Overview}}, 
      author={Haoxuan Qu and Hossein Rahmani and Li Xu and Bryan Williams and Jun Liu},
      year={2025},
      journal = {IET Computer Vision},
}

@inproceedings{zhou2023model,
  title={A Model or 603 Exemplars: Towards Memory-Efficient Class-Incremental Learning},
  author={Zhou, Da-Wei and Wang, Qi-Wei and Ye, Han-Jia and Zhan, De-Chuan},
  booktitle={ICLR},
  year={2023}
}
\newpage
\appendix
\section{Appendix}\label{appendix}
  The Appendix mainly contains additional materials and experiments that cannot be reported due to the page limit, which is organized as follows: 
\begin{itemize}
    \item  Section~\ref{RW} contains detailed Related Work of Continual Learning approaches.
    \item The Evaluation Protocol section~\ref{Evaluation Protocol} outlines the CL's primary scenarios used in this paper.
    \item The Evaluation Metrics section~\ref{Evalution Metrices} provides the mathematical definitions of the metrics used in this study.
    \item The "Things We Tried That Did Not Work" section~\ref{Things We Tried That Did Not Work} provides limitations and unsuccessful approaches that we took during the development of the method.
    \item The Additional Results section~\ref{Additional Results} presents the CIL and TIL scenario for CIFAR-100 and Tiny-ImageNet, task granularity, parameter usage efficiency, and ablation study. 
    \item The Pseudo-code section~\ref{Pseudo-code of the NFL and the NFL+} presents the algorithms for NFL and NFL+ in detail.
    \item The Implementation details of iCaRL section~\ref{Implementation Details of iCaRL} describe how the method was adapted for the TIL scenario, given that its original design specifically targets the CIL setting.
    \item The Hyperparameter section~\ref{Hyperparameter Search} provides details about the best hyperparameters selected, as well as a comprehensive list of all parameter combinations that were evaluated.

\end{itemize}
\section{Related Work}\label{RW} 
\textbf{Memory-based CL} methods maintain a buffer of past exemplars and interleave them with new task data during training. Incremental Classifier and Representation Learning (iCaRL)~\citep{rebuffi2017icarl} selects representative samples via herding and combines them with nearest-mean classification. Dark Experience Replay (DER++)~\citep{buzzega2020dark} stores both inputs and their corresponding logits to preserve inter-class relationships. Architecture-expanding methods offer an alternative: DyTox~\citep{Douillard_DyTox} grows a transformer-based model with task-specific tokens, and Memory-efficient Expandable MOdel (MEMO)~\citep{zhou2023model} expands deeper layers while sharing shallow features. Despite their effectiveness, memory-based methods that rely on replay buffers suffer from fundamental limitations: \textbf{i. Privacy Concerns} as storing raw samples may violate data protection regulations (e.g., GDPR~\citep{voigt2017gdpr}); \textbf{ii. Memory Overhead} since buffer size typically scales with the number of tasks, and \textbf{iii. Computational Cost} as replay mechanisms increase training time. These constraints substantially limit their applicability in scenarios with strict storage budgets or in cases where regulations prohibit exemplar storage.
 
\textbf{Buffer-free CL} methods operate without replay buffers and must rely on either regularization or structural constraints to prevent forgetting. Elastic Weight Consolidation (EWC)~\citep{kirkpatrick2017overcoming} penalizes changes to parameters deemed important by the Fisher Information Matrix (FIM), and Synaptic Intelligence (SI)~\citep{zenke2017continual} accumulates online importance estimates along the optimization trajectory. Learning without Forgetting (LwF)~\citep{li2017learningforgetting} distills knowledge from the previous model's predictions but degrades under distributional shift across tasks. More recently, SpaceNet~\citep{sokar2021spacenet} exploits network sparsity by freeing unused capacity for new tasks, Neuro-Inspired Stability-Plasticity Adaptation (NISPA)~\citep{gurbuz2022nispa} combines neurogenesis-inspired growth with stability-plasticity gating in sparse networks, Winning SubNetwork (WSN)~\citep{haeyong} targets Task Incremental Learning (TIL) by assigning binary masks to task-specific parameters, whereas Prediction Error-based Classification (PEC)~\citep{PEC} expands the network by introducing new modules per class and is limited to Class Incremental Learning (CIL). Discriminative and Consistent Network (DCNet)~\citep{wang2025dcnet} maps class representations into an orthogonal hyperspherical space to achieve inter-class separation without exemplars. While these methods eliminate the replay buffer, they typically introduce auxiliary mechanisms (sparsity masks, expanding modules, or fixed orthogonal embeddings) that constrain the model's flexibility. In contrast, the NFL operates entirely within the fixed capacity of a standard network, relying solely on the training schedule to prevent interference.
 
\textbf{Parameter-efficient CL} keeps the backbone frozen and adapts lightweight modules of large pre-trained models. Prompt-based methods such as Learning to Prompt (L2P)~\citep{L2p} and COntinual Decomposed Attention-based Prompting (CODA-Prompt)~\citep{Coda} condition the model through learned input tokens. LoRA-based methods constrain updates to low-rank subspaces: CL-LoRA~\citep{he2025cllora} assigns task-specific LoRA modules to isolate parameter updates, while EWC-LoRA~\citep{zheng2026revisiting} regularizes a shared LoRA module through the FIM estimated over the full-dimensional update space, achieving competitive performance with constant storage. Our NFL+LoRA builds on this direction by combining the stepwise-freezing philosophy of NFL with LoRA and Fisher regularization, and integrating knowledge distillation into the parameter-efficient regime.
\section{Evaluation Protocol}\label{Evaluation Protocol}

The two main experimental scenarios typically used to evaluate the performance of methods are the following: 
\begin{itemize}
    \item \textbf{Task Incremental Learning (TIL):} In TIL, the training data is divided into multiple tasks, each with a unique set of classes. The crucial aspect of TIL is that the model is provided with information about which task it is handling during training and testing. This allows the model to use the computational graph corresponding to each task. For example, if the model is trained to classify images of animals and vehicles, the task label information is also provided for testing on a new image; thus, the network's classification output for the corresponding task will be calculated. This knowledge simplifies the inference task, as the model does not need to consider all possible classes simultaneously ~\citep{wickramasinghe2024continual, vahedifar_2025_14631802}.

    \item \textbf{Class Incremental Learning (CIL):} In CIL, the model is also trained on different tasks, but is not told which task a new sample belongs to during testing. Instead, regardless of the task, the model needs to respond to all the classes it has encountered. This makes CIL more challenging than TIL, as the model must infer the correct class without task-related information. For instance, after training a model to recognize animals and vehicles separately, CIL would test the model on all classes simultaneously (animals and vehicles) without informing the model whether it is currently classifying an animal or a vehicle ~\citep{qu2024recentadvancescontinuallearning, zhou2024continuallearningpretrainedmodels}.

\end{itemize}
Physically, our model utilizes a single output layer matrix $W_{\text{out}} \in \mathbb{R}^{(C_{\text{old}} + C_{\text{new}}) \times \mathrm{P}}$, where $\mathrm{P}$ is the feature dimension. However, logically, we treat this matrix as a concatenation of two blocks: $\theta_t$ (columns corresponding to previously learned classes) and $\theta_{t+1}$ (columns corresponding to new classes). The breakdown of training into steps relies on masking specific gradients rather than separating architectures. In the last step the entire single head is fine-tuned jointly, but the loss function remains logically split: $\mathcal{L}_{\text{KD}}$ is applied to the logits indexed by the old task, and $\mathcal{L}_{\text{CE}}$ is applied to the logits indexed by the new task.
Consequently, task identifiers are strictly a \textit{training-time} requirement used to define the boundaries of these logical partitions for loss computation. During inference, particularly in the CIL setting, the partitions are ignored, and the single head functions as a global classifier over the union of all classes $\mathcal{C}_t \cup \mathcal{C}_{t+1}$.

\section{Evaluation Metrics}\label{Evalution Metrices}
When CL methods trained for task $T_k$ on $\mathcal{D}_k$, its accuracy on all tasks is measured using the corresponding test sets, leading to a matrix $A \in \mathbb{R}^{T \times T}$ containing the accuracies on all $T$ tasks, i.e., $A_{i,j}$ denotes the accuracy of the model on task $T_j$ after trained completely on task $T_i$.

 \textbf{Average Accuracy (ACC)}~\citep{Lopez-PazNeurIPS2017_f8752278}:  This metric assesses the overall performance of the CL method after completing the training of all $T$ tasks.
    \begin{equation}
    \text{ACC}_T = \frac{1}{T} \sum_{k=1}^{T} A_{T,k}.
    \end{equation}
      
 \textbf{Backward transfer (BWT)}~\citep{Lopez-PazNeurIPS2017_f8752278}: This metric evaluates the average influence of learning the $T$-th task on previous tasks:
    \begin{equation}
    \text{BWT}_T = \frac{1}{T-1} \sum_{k=1}^{T-1} \left(A_{T,k} - A_{k,k}\right).
    \end{equation}  
In summary, for a CL method, the higher the ACC, BWT and PS in the trained models, the better the method is to combat catastrophic Forgetting~\citep{diazrodriguez2018dontforgetforgettingnew}. Notably, Backward transfer for the first task is meaningless~\citep{Lopez-PazNeurIPS2017_f8752278}.
\section{Limitations: "Things We Tried That Did Not Work"}\label{Things We Tried That Did Not Work}

We document several directions we explored that did not yield the expected improvements, as well as aspects of the framework that remain unresolved. 

\paragraph{1. Sensitivity to task ordering.}
NFL assumes a fixed, arbitrary task ordering. We experimented with curriculum-based orderings (easy-to-hard, similarity-based) to determine whether a principled task sequence could improve performance. The results were inconclusive: gains on one dataset were offset by degradation on another. The framework currently lacks a mechanism to adapt its freezing schedule to the relationship between consecutive tasks, and performance can vary by 2--4\% in ACC depending on the task permutation.

\paragraph{2. Scalability beyond binary task transitions.}
The stepwise freezing pipeline is designed around a two-task transition ($T_t \rightarrow T_{t+1}$). When many tasks arrive in rapid succession, the multi-step overhead per transition accumulates. We attempted to amortize this cost by batching multiple tasks into a single transition cycle, but this reintroduced the forgetting that the stepwise isolation is designed to prevent.

\paragraph{3. Long task sequences and Fisher accumulation drift.}
In NFL+LoRA, the accumulated Fisher matrix $\mathbf{F}^{\text{cum}}_t$ is intended to capture parameter importance across all previous tasks. On sequences beyond 20 tasks, we observed that the Fisher estimates become increasingly stale: the cosine similarity between $\mathbf{F}^{\text{cum}}_t$ and the true task-specific Fisher decays noticeably. We experimented with rehearsal-based Fisher re-estimation, but this contradicts the memory-free constraint. The decay factor $\gamma$ provides partial mitigation, yet there is no principled way to set it without validation data.

\paragraph{4. Auto-encoder bottleneck dimension.}
In NFL+, the under-complete auto-encoder requires selecting a bottleneck dimension that balances feature compression with reconstruction fidelity. We found this hyperparameter to be dataset-sensitive: the optimal bottleneck for CIFAR-100 performed poorly on ImageNet-1000 and vice versa. We attempted adaptive bottleneck sizing based on the singular value spectrum of the feature covariance matrix, but the added complexity did not yield consistent gains.

\paragraph{5. LoRA rank selection across tasks.}
NFL+LoRA uses a fixed LoRA rank $r$ for all tasks. We experimented with adaptive rank allocation, assigning a higher rank to tasks that require greater plasticity (measured by the initial training loss). While this improved plasticity on harder tasks, it introduced instability in Fisher accumulation because the dimensionality of the update space varied across tasks. A principled rank-scheduling strategy compatible with cumulative Fisher estimation remains an open problem.

\paragraph{6. Bias correction without the auto-encoder.}
NFL+ employs the auto-encoder's encoder to construct a logit transformation $\Gamma$ that corrects prediction bias toward new classes. When transitioning to NFL+LoRA, which removes the auto-encoder, we attempted alternative bias-correction strategies, including weight normalization of the classification head and post hoc calibration via temperature scaling, but none matched the effectiveness of the learned transformation in NFL+.

\paragraph{7. Computational cost of the multi-step pipeline.}
Each task transition in the NFL requires four training phases. While each individual step is lightweight, the cumulative wall-clock time per task is approximately 2.5--3$\times$ that of a single fine-tuning pass. We explored distilling the multi-step procedure into a single training phase with a composite loss (combining all distillation and regularization terms simultaneously), but this collapsed the stability--plasticity separation that the sequential stages enforce, resulting in performance comparable to standard LwF.

\paragraph{8. Generalization to non-vision modalities.}
All experiments are conducted on image classification benchmarks. We did not evaluate the NFL on language or multi-modal CL tasks. The stepwise freezing strategy is architecture-agnostic in principle, but the specific design choices, task-specific classification heads, feature-level auto-encoding, LoRA placement at attention layers, are tailored to vision architectures. Whether the same freezing sequence is effective for both decoder-only language models and sequence-to-sequence tasks remains unknown. We will investigate this in future work.

\paragraph{9. Interaction between knowledge distillation and Fisher regularization.}
In NFL+LoRA, the loss function (Eq.~\ref{eq:nflpp_loss}) combines distillation terms with a Fisher-weighted quadratic penalty. We observed that these two mechanisms can conflict: distillation encourages the model to match previous soft targets (a functional constraint), while Fisher regularization penalizes parameter movement (a structural constraint). At high $\lambda$, the Fisher term dominates and the distillation signal is effectively suppressed. We did not find a principled way to jointly calibrate $\lambda$ and the distillation weight $\alpha$; in practice, we tune them independently via grid search.

\paragraph{10. Lack of theoretical guarantees.}
The stepwise freezing schedule is motivated by an exhaustive analysis of the $2^3 = 8$ training configurations, though it is empirical. We do not provide formal proof of the optimality of the selected configuration sequence.

\section{Additional Results}\label{Additional Results}

In this section, we present the following results:
\begin{enumerate}
    \item The CIL scenario results for CIFAR-100 and Tiny-ImageNet datasets in Section~\ref{section CIL CifarTiny}. (See Table~\ref{tab:CIL_cifar_tiny}). 
    \item The TIL scenario results for CIFAR-100 and Tiny-ImageNet datasets in Section~\ref{Section TIL cifarTiny}. (See Table~\ref{tab:TIL_cifar_tiny}).
    \item Analysis of the NFL+ accuracy matrix in Section~\ref{Analysis of the NFL+}. (See Figs.\ref{fig: 10task_NFL+_matrix} and \ref{fig: 20task_NFL+_matrix}). 
    \item Task Granularity Analysis for CIL in section~\ref{Task Granularity Analysis for CIL}. (See Fig.~\ref{fig:CIL20tasks})
    \item Memory-wise comparison for CIFAR-100 in section~\ref{section Memory-wise comparison}. (See Table~\ref{tab:memory_usage}).
    \item Computation cost for all datasets, measuring the FLOPs, training and inference time, and GPU peak time provided in section~\ref{section Computational Cost Comparison}. (See Table~\ref{tab:compute_all}).
    \item Parameter Usage Efficiency Analysis in section~\ref{Parameter Usage Efficiency}. (See Fig.~\ref{fig:weight_pruning}).
    \item  An ablation study of each step's contribution for the NFL and the NFL+ in section~\ref{Ablation Study of the NFL and the NFL+}. (See Table~\ref{tab:NFL_ablation} and Table~\ref{tab:NFL_plus_ablation})
\end{enumerate}

\subsection{Comparison based on the CIL Scenario}\label{section CIL CifarTiny} 
\begin{table*}[t]
\centering
\caption{CIFAR-100 and Tiny-ImageNet results for the CIL scenario. Best results are highlighted for \textcolor{green!80}{buffer-free} and \textcolor{cyan!70}{memory-based}.}
\label{tab:CIL_cifar_tiny}
\vspace{0.5em}
\small
\setlength{\tabcolsep}{3.5pt}
\renewcommand{\arraystretch}{1}
\resizebox{0.95\textwidth}{!}{\begin{tabular}{@{}l ccc ccc ccc ccc@{}}
\toprule
& \multicolumn{6}{c}{\textbf{CIFAR-100}} & \multicolumn{6}{c}{\textbf{Tiny-ImageNet}} \\
\cmidrule(lr){2-7} \cmidrule(lr){8-13}
& \multicolumn{3}{c}{\textbf{10 tasks}} & \multicolumn{3}{c}{\textbf{20 tasks}} & \multicolumn{3}{c}{\textbf{10 tasks}} & \multicolumn{3}{c}{\textbf{20 tasks}} \\
\cmidrule(lr){2-4} \cmidrule(lr){5-7} \cmidrule(lr){8-10} \cmidrule(lr){11-13}
\textbf{Method} & ACC$\uparrow$ & BWT$\uparrow$ & PS$\uparrow$ & ACC$\uparrow$ & BWT$\uparrow$ & PS$\uparrow$ & ACC$\uparrow$ & BWT$\uparrow$ & PS$\uparrow$ & ACC$\uparrow$ & BWT$\uparrow$ & PS$\uparrow$ \\
\midrule
\multicolumn{13}{c}{\textit{\textbf{Memory-free methods}}} \\
NFL+ (Ours)
  & \cellcolor{green!20}53.70 \scriptsize{±2.18} & $-$38.73 & 0.70
  & \cellcolor{green!20}44.03 \scriptsize{±2.85} & $-$42.60 & \cellcolor{green!20}0.60
  & \cellcolor{green!20}44.70 \scriptsize{±3.12} & \cellcolor{green!20}$-$40.26 & \cellcolor{green!20}0.68
  & \cellcolor{green!20}36.65 \scriptsize{±3.41} & \cellcolor{green!20}$-$44.29 & \cellcolor{green!20}0.56 \\
DCNet
  & 52.85 \scriptsize{±2.25} & \cellcolor{green!20}$-$38.18 & \cellcolor{green!20}0.72
  & 43.10 \scriptsize{±2.92} & \cellcolor{green!20}$-$42.00 & 0.58
  & 43.90 \scriptsize{±3.18} & $-$41.28 & \cellcolor{green!20}0.68
  & 35.80 \scriptsize{±3.48} & $-$45.41 & 0.54 \\
NISPA
  & 43.80 \scriptsize{±2.95} & $-$46.05 & 0.66
  & 35.20 \scriptsize{±3.40} & $-$49.73 & 0.51
  & 35.40 \scriptsize{±3.65} & $-$45.48 & 0.63
  & 27.90 \scriptsize{±3.92} & $-$49.12 & 0.47 \\
NFL
  & 41.20 \scriptsize{±3.45} & $-$44.80 & 0.59
  & 31.85 \scriptsize{±4.12} & $-$48.38 & 0.48
  & 32.50 \scriptsize{±4.28} & $-$48.50 & 0.50
  & 24.80 \scriptsize{±4.65} & $-$52.38 & 0.44 \\
PEC
  & 29.40 \scriptsize{±2.87} & $-$53.04 & 0.58
  & 24.11 \scriptsize{±3.24} & $-$56.22 & 0.48
  & 19.40 \scriptsize{±3.56} & $-$52.96 & 0.54
  & 15.91 \scriptsize{±3.89} & $-$56.14 & 0.42 \\
SpaceNet
  & 27.10 \scriptsize{±3.15} & $-$51.07 & 0.57
  & 21.80 \scriptsize{±3.48} & $-$54.64 & 0.40
  & 17.80 \scriptsize{±3.78} & $-$50.31 & 0.54
  & 14.20 \scriptsize{±4.05} & $-$53.83 & 0.36 \\
LwF
  & 19.56 \scriptsize{±3.12} & $-$52.81 & 0.54
  & 16.04 \scriptsize{±3.45} & $-$55.98 & 0.38
  & 15.56 \scriptsize{±3.78} & $-$51.50 & 0.52
  & 12.76 \scriptsize{±4.02} & $-$54.59 & 0.39 \\
SI
  & 15.37 \scriptsize{±4.25} & $-$51.13 & 0.52
  & 13.23 \scriptsize{±4.56} & $-$54.71 & 0.41
  & 13.37 \scriptsize{±4.87} & $-$51.84 & 0.51
  & 12.02 \scriptsize{±5.12} & $-$55.47 & 0.43 \\
EWC
  & 12.87 \scriptsize{±3.56} & $-$46.99 & 0.51
  & 10.55 \scriptsize{±3.89} & $-$50.75 & 0.37
  & 10.87 \scriptsize{±4.12} & $-$49.57 & 0.50
  & 8.91 \scriptsize{±4.35} & $-$53.54 & 0.40 \\
\midrule
\multicolumn{13}{c}{\textit{\textbf{Memory-based methods}}} \\
DyTox
  & \cellcolor{cyan!20}55.80 \scriptsize{±2.52} & $-$30.50 & 0.71
  & \cellcolor{cyan!20}45.50 \scriptsize{±3.20} & $-$33.85 & 0.62
  & \cellcolor{cyan!20}47.60 \scriptsize{±3.35} & $-$32.80 & 0.67
  & \cellcolor{cyan!20}38.80 \scriptsize{±3.72} & $-$36.42 & 0.58 \\
MEMO
  & 54.40 \scriptsize{±2.85} & $-$31.80 & \cellcolor{cyan!20}0.78
  & 44.10 \scriptsize{±3.42} & $-$35.20 & \cellcolor{cyan!20}0.68
  & 46.20 \scriptsize{±3.58} & $-$34.50 & \cellcolor{cyan!20}0.74
  & 37.40 \scriptsize{±3.95} & $-$38.15 & \cellcolor{cyan!20}0.63 \\
DER++
  & 21.20 \scriptsize{±3.48} & \cellcolor{cyan!20}$-$28.20 & 0.38
  & 17.10 \scriptsize{±3.72} & \cellcolor{cyan!20}$-$31.50 & 0.35
  & 18.10 \scriptsize{±4.05} & \cellcolor{cyan!20}$-$30.40 & 0.40
  & 14.60 \scriptsize{±4.30} & \cellcolor{cyan!20}$-$33.75 & 0.36 \\
iCaRL
  & 21.50 \scriptsize{±4.35} & $-$52.40 & 0.39
  & 17.30 \scriptsize{±4.65} & $-$56.10 & 0.35
  & 15.80 \scriptsize{±4.95} & $-$54.80 & 0.36
  & 12.50 \scriptsize{±5.18} & $-$58.60 & 0.32 \\
\bottomrule
\end{tabular}
}
\end{table*}
Table~\ref{tab:CIL_cifar_tiny} summarizes performance in the CIL setting across CIFAR-100 and Tiny-ImageNet. NFL+ consistently demonstrates strong results, achieving 53.70\% accuracy on CIFAR-100, closely matching DyTox (55.80\%) and MEMO (54.40\%), both of which rely on significantly larger models or memory footprints. Similar trends are observed on Tiny-ImageNet and ImageNet-1000, where NFL+ remains competitive with DyTox and MEMO. Notably, NFL+ outperforms most memory-based methods, surpassing iCaRL (21.50\%) and DER++ (21.20\%) by more than 32\% improvement on CIFAR-100. Even the base NFL model, without Auto-Encoder, achieves 41.20\% on CIFAR-100 and 32.50\% on Tiny-ImageNet, still surpassing iCaRL and DER++ by wide margins. These results indicate that a well-designed buffer-free method can exhibit greater structural stability than memory-based methods that rely on underutilized buffers. Additionally, among buffer-free approaches, NFL+ establishes a clear lead, outperforming PEC, the next-best method, across three datasets. Another notable observation is that the performance of all methods declines noticeably as the number of tasks increases. For instance, on CIFAR-100, NFL+ sees a drop in accuracy from 53.70\% to 44.03\% as the number of tasks increases from 10 to 20, suggesting that an increasing task count is a significant challenge for CL methods.   Furthermore, NFL+ exhibits substantially lower forgetting, achieving a BWT of $-0.05$ compared to BWTs of $-0.42$ and $-0.44$ for DyTox and MEMO, respectively, indicating that its performance gains do not come at the cost of stability.

\subsection{Comparison based on the TIL Scenario}\label{Section TIL cifarTiny}
As shown in Table~\ref{tab:TIL_cifar_tiny}, performance trends remain consistent across datasets comparing the TIL to the CIL. The TIL scenario is inherently easier, as task boundaries are known during inference and models only classify among classes within the current task. This is reflected in the accuracy gap: NFL+ achieves 70.68\% on CIFAR-100 in the TIL compared to 53.70\% in the CIL, a difference of 17\% improvement. Similarly, on ImageNet-1000, NFL+ reaches 51.36\% in the TIL compared to 38.42\% in the CIL. The same pattern holds for memory-based methods: DyTox achieves 73.80\% (TIL) compared to 55.80\% (CIL) on CIFAR-100, confirming that CIL's requirement to distinguish among all previously learned classes without task identity poses a substantially greater challenge.

By analyzing Table~\ref{tab:TIL_cifar_tiny}, we can observe that the ACC across all methods tends to decrease slightly when moving from 10 to 20 tasks. This is expected as increasing the number of incremental steps generally increases task complexity. However, the NFL+ consistently achieves the highest ACC in both settings (70.68 for 10 tasks in CIFAR-100 and 62.14 for 20 tasks in CIFAR-100), demonstrating its effectiveness. The NFL+ maintains the highest plasticity in both cases (0.72 and 0.65 for 10 and 20 tasks, respectively). The performance trends across most methods align with those observed on CIFAR-100. However, on large-scale datasets, the performance of these methods degrades as the number of classes to be learned per task increases. These results suggest that when designing a CL framework, ACC should not be the sole evaluation metric. The PS metric is equally important, as it reflects the model’s ability to acquire new knowledge (plasticity) while retaining previously learned information (stability). Notably, NFL+ narrows the gap with state-of-the-art methods more effectively in TIL, achieving 70.68\% compare to 73.80\% of DyTox's accuracy on CIFAR-100. Furthermore, NFL+ exhibits dramatically lower forgetting than traditional regularization methods: on CIFAR-100 (10 tasks), NFL+ achieves a BWT of -0.35 compared to LwF (-39.69), SI (-45.78), and EWC(-31.64), a reduction of over 98\%.
\begin{table*}[t]
\centering
\caption{ACC, BWT, and PS performance comparison for TIL scenario on CIFAR-100 and Tiny-ImageNet. Best results are highlighted for \textcolor{green!80}{buffer-free} and \textcolor{cyan!70}{memory-based}. Memory-based methods use a fixed buffer of 200 exemplars.}
\label{tab:TIL_cifar_tiny}
\setlength{\tabcolsep}{2pt}
\renewcommand{\arraystretch}{1}
\resizebox{.95\textwidth}{!}{\begin{tabular}{@{}l ccc ccc ccc ccc@{}}
\toprule
& \multicolumn{6}{c}{\textbf{CIFAR-100}} & \multicolumn{6}{c}{\textbf{Tiny-ImageNet}} \\
\cmidrule(lr){2-7} \cmidrule(lr){8-13}
& \multicolumn{3}{c}{\textbf{10 tasks}} & \multicolumn{3}{c}{\textbf{20 tasks}} & \multicolumn{3}{c}{\textbf{10 tasks}} & \multicolumn{3}{c}{\textbf{20 tasks}} \\
\cmidrule(lr){2-4} \cmidrule(lr){5-7} \cmidrule(lr){8-10} \cmidrule(lr){11-13}
\textbf{Method} & ACC$\uparrow$ & BWT$\uparrow$ & PS$\uparrow$ & ACC$\uparrow$ & BWT$\uparrow$ & PS$\uparrow$ & ACC$\uparrow$ & BWT$\uparrow$ & PS$\uparrow$ & ACC$\uparrow$ & BWT$\uparrow$ & PS$\uparrow$ \\
\midrule
\multicolumn{13}{c}{\textit{\textbf{Memory-free methods}}} \\
NFL+ (Ours)
  & \cellcolor{green!20}70.68 \scriptsize{±2.45} & $-$0.35 & \cellcolor{green!20}0.72
  & \cellcolor{green!20}62.14 \scriptsize{±3.12} & $-$0.52 & \cellcolor{green!20}0.65
  & \cellcolor{green!20}58.21 \scriptsize{±4.10} & $-$1.04 & \cellcolor{green!20}0.68
  & \cellcolor{green!20}49.85 \scriptsize{±3.87} & $-$1.35 & \cellcolor{green!20}0.61 \\
DCNet
  & 66.80 \scriptsize{±2.70} & $-$1.68 & 0.66
  & 58.50 \scriptsize{±3.35} & $-$2.30 & 0.59
  & 54.80 \scriptsize{±4.32} & $-$2.90 & 0.62
  & 46.20 \scriptsize{±4.02} & $-$3.75 & 0.54 \\
WSN
  & 64.00 \scriptsize{±0.36} & \cellcolor{green!20}0.0 & 0.66
  & 55.82 \scriptsize{±0.52} & \cellcolor{green!20}0.0 & 0.58
  & 61.06 \scriptsize{±1.02} & \cellcolor{green!20}0.0 & 0.64
  & 52.41 \scriptsize{±1.38} & \cellcolor{green!20}0.0 & 0.55 \\
NISPA
  & 62.35 \scriptsize{±3.10} & $-$3.48 & 0.62
  & 54.10 \scriptsize{±3.65} & $-$5.20 & 0.54
  & 53.80 \scriptsize{±4.42} & $-$5.65 & 0.59
  & 45.20 \scriptsize{±4.18} & $-$8.10 & 0.51 \\
NFL
  & 58.42 \scriptsize{±3.18} & $-$9.22 & 0.58
  & 49.37 \scriptsize{±4.21} & $-$11.85 & 0.51
  & 47.65 \scriptsize{±5.13} & $-$14.73 & 0.54
  & 38.92 \scriptsize{±4.56} & $-$18.21 & 0.47 \\
SpaceNet
  & 54.15 \scriptsize{±3.72} & $-$12.40 & 0.48
  & 45.50 \scriptsize{±4.08} & $-$16.80 & 0.41
  & 48.90 \scriptsize{±4.50} & $-$15.35 & 0.46
  & 40.25 \scriptsize{±4.22} & $-$20.70 & 0.39 \\
LwF
  & 52.86 \scriptsize{±3.50} & $-$39.69 & 0.45
  & 44.21 \scriptsize{±4.15} & $-$43.52 & 0.39
  & 49.85 \scriptsize{±1.59} & $-$41.21 & 0.43
  & 41.37 \scriptsize{±2.84} & $-$46.18 & 0.37 \\
SI
  & 51.13 \scriptsize{±6.25} & $-$45.78 & 0.47
  & 42.68 \scriptsize{±5.73} & $-$49.31 & 0.41
  & 48.45 \scriptsize{±4.87} & $-$52.97 & 0.49
  & 40.12 \scriptsize{±5.21} & $-$57.64 & 0.43 \\
EWC
  & 50.59 \scriptsize{±1.39} & $-$31.64 & 0.44
  & 41.85 \scriptsize{±2.46} & $-$36.21 & 0.38
  & 44.19 \scriptsize{±3.45} & $-$41.85 & 0.46
  & 35.74 \scriptsize{±4.12} & $-$47.32 & 0.40 \\
\midrule
\multicolumn{13}{c}{\textit{\textbf{Memory-based methods}}} \\
DyTox
  & \cellcolor{cyan!20}73.80 \scriptsize{±2.52} & $-$3.50 & 0.69
  & \cellcolor{cyan!20}65.60 \scriptsize{±3.22} & $-$4.45 & 0.61
  & \cellcolor{cyan!20}68.50 \scriptsize{±3.28} & $-$3.55 & 0.66
  & \cellcolor{cyan!20}59.30 \scriptsize{±3.62} & $-$4.75 & 0.57 \\
MEMO
  & 73.00 \scriptsize{±2.75} & \cellcolor{cyan!20}$-$2.65 & \cellcolor{cyan!20}0.74
  & 64.80 \scriptsize{±3.35} & \cellcolor{cyan!20}$-$3.50 & \cellcolor{cyan!20}0.67
  & 67.60 \scriptsize{±3.52} & \cellcolor{cyan!20}$-$2.80 & \cellcolor{cyan!20}0.71
  & 58.40 \scriptsize{±3.85} & \cellcolor{cyan!20}$-$3.90 & \cellcolor{cyan!20}0.63 \\
DER++
  & 70.60 \scriptsize{±2.72} & $-$6.50 & 0.40
  & 62.00 \scriptsize{±3.50} & $-$7.50 & 0.34
  & 67.80 \scriptsize{±3.72} & $-$3.55 & 0.43
  & 59.00 \scriptsize{±4.15} & $-$4.85 & 0.37 \\
iCaRL
  & 70.10 \scriptsize{±4.08} & $-$5.35 & 0.43
  & 61.40 \scriptsize{±4.42} & $-$6.55 & 0.37
  & 66.70 \scriptsize{±3.02} & $-$6.45 & 0.41
  & 58.10 \scriptsize{±4.00} & $-$7.85 & 0.35 \\
\bottomrule
\end{tabular}
}
\end{table*}
\subsection{Analysis of the NFL+ Accuracy Matrix}\label{Analysis of the NFL+}
\begin{figure*}[t]
    \centering
    \includegraphics[width=\linewidth]{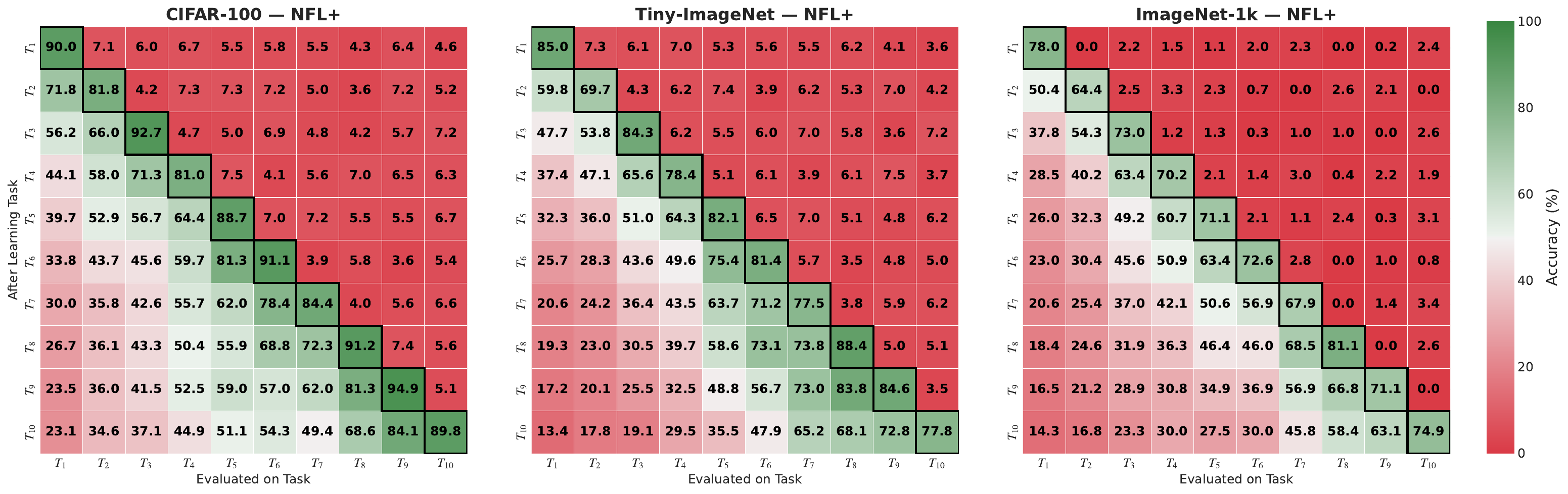}
    \caption{ACC matrix for NFL+ for the CIL for 10 tasks for each dataset.}
    \label{fig: 10task_NFL+_matrix}
\end{figure*}
\begin{figure*}[t]
    \centering
    \includegraphics[width=\linewidth]{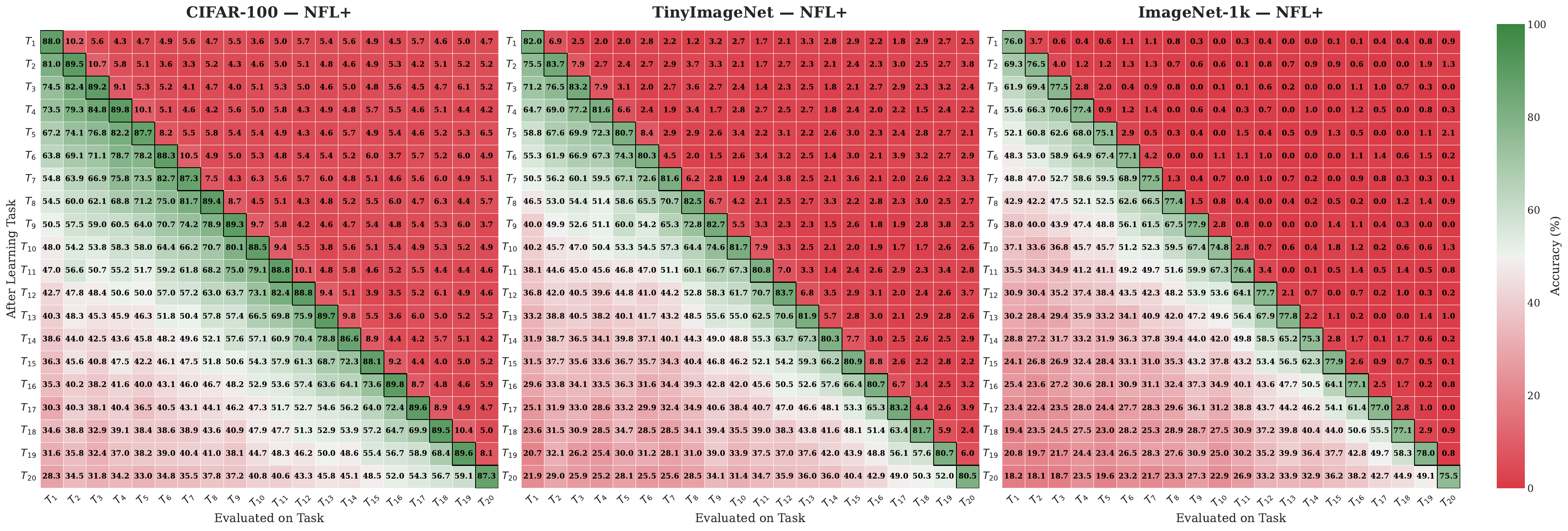}
    \caption{ACC matrix for NFL+ for the CIL for 20 tasks for each dataset.}
    \label{fig: 20task_NFL+_matrix}
\end{figure*}
Examining the NFL+ matrices across datasets reveals several characteristic patterns, as shown in Fig.~\ref{fig: 10task_NFL+_matrix} and Fig.~\ref{fig: 20task_NFL+_matrix}. The diagonal entries remain consistently high across all tasks, ranging from approximately 85-92\% on CIFAR-100, 80-88\% on Tiny-ImageNet, and 72-80\% on ImageNet-1000. This uniformity indicates that NFL+ maintains strong performance on individual tasks.
Additionally, earlier tasks (those learned first) exhibit greater forgetting than more recently learned tasks, consistent with the temporal dynamics of CL. However, the magnitude of forgetting in NFL+ is notably smaller than in regularization-based methods. For instance, on CIFAR-100 with 10 tasks, the first task drops from approximately 90\% to around 45\%, representing roughly 45\% forgetting. While this forgetting is non-trivial, it compares favorably with methods like EWC and SI, in which earlier tasks can degrade to near-random performance.

Comparing the 10-task in Fig.~\ref{fig: 10task_NFL+_matrix} with 20-task in Fig,~\ref{fig: 20task_NFL+_matrix} reveals that the forgetting rate per task remains relatively constant, but the cumulative effect is more pronounced with more tasks. In the 20-task scenario, the first task must survive 19 subsequent learning episodes, resulting in greater total forgetting compared to the 10-task case, where only 9 subsequent tasks are learned.

\begin{figure*}[t]
    \centering
    \includegraphics[width=\textwidth]{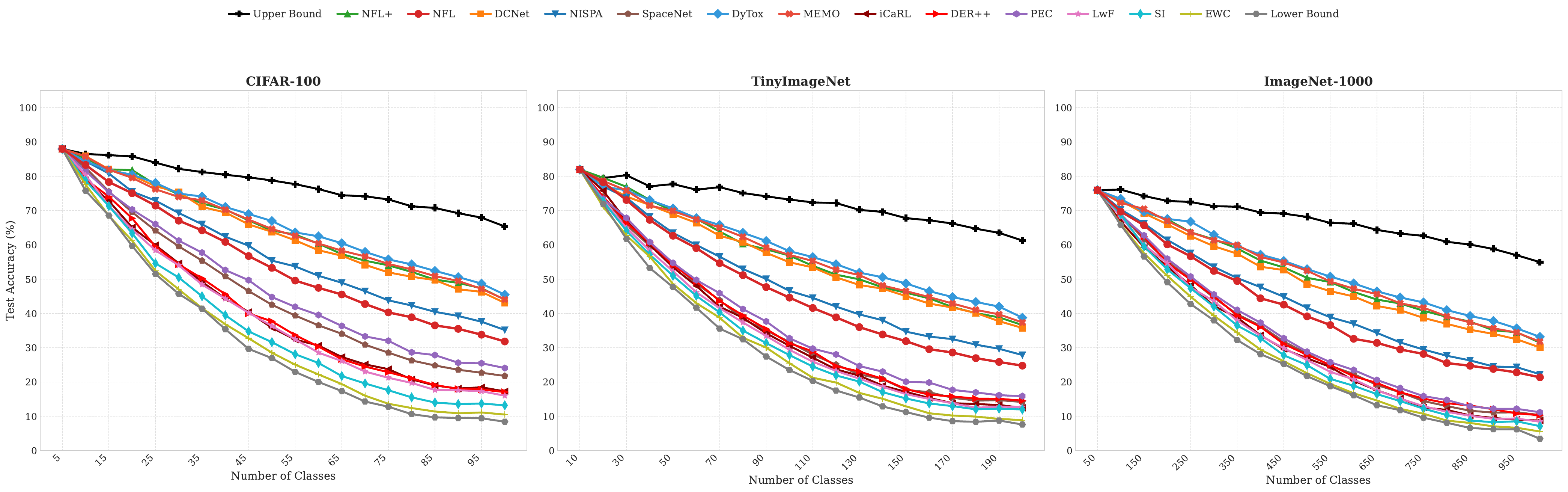}
    \caption{ACC evaluation comparison for the CIL across 20 tasks per dataset. Each point represents the average classification accuracy evaluated after learning a given task, averaged over all tasks learned up to that point. For example, the value for task 15 corresponds to the average accuracy of the model on the test sets for tasks 1 through 15 after training on task 15. For details, see Fig.~\ref{fig: 20task_NFL+_matrix}. Solid lines: memory-based; dashed lines: buffer-free.}
    \label{fig:CIL20tasks}
\end{figure*}
\subsection{Task Granularity Analysis for CIL}\label{Task Granularity Analysis for CIL}
We report the results of the CIL scenario for each dataset under a configuration of 20 tasks, as illustrated in Fig.~\ref{fig:CIL20tasks}. In this setting, the 100, 200, and 1000 classes from CIFAR-100, Tiny-ImageNet, and ImageNet-1000 correspond to 5, 10, and 50 incremental classes per task, respectively. This setup allows us to analyze how both the number of tasks and the number of classes per task influence overall performance. Fig.~\ref{fig:CIL20tasks} demonstrates that performance consistently declines as the number of tasks increases. This observation raises an important question: \textit{Given a fixed number of classes, what is the optimal task partitioning strategy to maintain the best performance?}

The mapping results in Fig.~\ref{fig:CIL10task} and Fig.~\ref{fig:CIL20tasks} reveal that all methods experience performance degradation when the number of tasks increases from 10 to 20, which is expected, since finer task granularity introduces more learning episodes and thus more opportunities for Catastrophic Forgetting. On CIFAR-100, NFL+ decreases from 53.70\% to 44.03\% (9.67\% decreases). Interestingly, the performance gap between NFL+ and the leading memory-based method, DyTox, actually narrows slightly in the 20-task scenario, from 2.10\% to 1.47\% on CIFAR-100 and from 2.90\% to 2.15\% on Tiny-ImageNet. The evolution curves show that NFL+ exhibits a more gradual decline in accuracy than DyTox. While methods like EWC and SI show steep initial drops followed by plateaus at low accuracy, NFL+ maintains a smoother degradation trajectory that tracks more closely with memory-based methods. 

Furthermore, the granularity of task division significantly affects ACC, especially as the number of tasks grows. This suggests that many existing CL methods implicitly rely on grouping many classes into a single task. Consequently, the feature extractor becomes critical, since learning multiple classes simultaneously requires extracting a richer set of features. In addition, we observe bias in the final classification layer. Specifically, the last layer tends to favor newly introduced classes, as it is updated with their data, which in turn makes it less effective at preserving knowledge of previously learned classes. These findings suggest the need for buffer-free methods that more effectively coordinate the feature extractor and classification head.

\subsection{Memory-wise comparison}\label{section Memory-wise comparison}
We evaluate the total memory cost of ResNet-18 on ImageNet-1000 by summing the memory required for storing exemplars and model parameters. Each ImageNet-1000 exemplar requires \(3 \times 224 \times 224 = 150,528\) bytes. With [20,000] exemplars, the total memory footprint for stored images is approximately \(3,010.56 \, \text{MB}\) across exemplar-based methods. The model memory varies based on the number of model parameters. Each parameter is stored as a 32-bit float (4 bytes). Most methods, including LWF, iCaRL, DER++, and NFL, contain 11.17 Million~(M) parameters, hence requiring \(44.68 \, \text{MB}\), while NFL+ has 11.56 M parameters (i.e. \(46.24 \, \text{MB}\)). DyTox is the most memory-intensive, requiring \(1832.51 \, \text{MB}\).
\begin{table}[t]
\centering
\caption{Memory footprint comparison across CL methods (in MB) for ImageNet-1000. The highest memory usage results are highlighted for for \textcolor{green!80}{buffer-free} and \textcolor{cyan!70}{memory-based}.}
\label{tab:memory_usage}
\vspace{0.5em}
\small
\setlength{\tabcolsep}{5pt}
\renewcommand{\arraystretch}{1}
\begin{tabular}{@{}l ccc@{}}
\toprule
\textbf{Method} & \textbf{Model Size} & \textbf{Buffer Size} & \textbf{Total} \\
\midrule
\multicolumn{4}{c}{\textit{\textbf{Memory-free methods}}} \\
NFL   & 44.7 & 0.0& 44.7 \\
NFL+  & \cellcolor{green!10}46.2 & 0.0& \cellcolor{green!10}46.2 \\
EWC   & 44.7 & 0.0& 44.7 \\
SI    & 44.7 & 0.0& 44.7 \\
LwF   & 44.7 & 0.0& 44.7 \\
\midrule
\multicolumn{4}{c}{\textit{\textbf{Memory-based methods}}} \\
iCaRL & 44.7 & 3010.6 & 3055.2 \\
DER++ & 44.7 & 3010.6 & 3055.2 \\
MEMO  & 682.4 & 3010.6 & 3693.0 \\
DyTox & \cellcolor{cyan!20}1832.5 & 3010.6 & \cellcolor{cyan!20}4843.1 \\
\bottomrule
\end{tabular}
\end{table}
As shown in Table~\ref{tab:memory_usage}, NFL+ achieves substantial improvements compared to DyTox, requiring only \textbf{2.53\%} of DyTox's \textbf{model size}. This contrast underscores a fundamental issue in CL evaluation: \textit{The unfair comparison of methods with vastly different memory footprints}.  This is primarily because memory-based methods and dynamic architectures inherently allocate additional memory and model resources, providing a clear advantage over buffer-free methods, which do not utilize exemplar buffers. 

The growing model size in dynamic-architecture methods, such as DyTox and MEMO, challenges the validity of current evaluation frameworks. Additionally, raises significant concerns about computational scalability and practical deployment in resource-constrained environments. To mitigate this unfair comparison, the authors in~\citep{zhou2024class} proposed increasing the exemplar capacity for memory-based approaches. However, this adjustment is fundamentally incompatible with buffer-free approaches (LwF, EWC, SI, NFL, NFL+), which by design do not store exemplars. Moreover, even among memory-based methods, simply expanding buffers cannot compensate for the architectural advantages of dynamic models. These findings underscore the need for evaluation frameworks that explicitly account for memory consumption, ensuring that reported performance gains reflect algorithmic innovation rather than resource asymmetry.

\subsection{Computational Cost Comparison}\label{section Computational Cost Comparison}
As shown in Table~\ref{tab:compute_all}, among all memory-free methods, NFL+ achieves the highest accuracy at moderate compute, matching MEMO's training budget while requiring no replay buffer, up to 5.6 GB less GPU memory, and 3$\times$ faster inference. Competitive baselines DCNet and NISPA demand comparable or higher FLOPs for lower accuracy; NISPA's progressive architecture search incurs 1.3$\times$ the training cost of NFL+ on ImageNet-1000.

\begin{table*}[t]
\caption{Computational cost across all datasets (CIL, 10 tasks,
single NVIDIA A6000 GPU). Training time is wall-clock for the
complete CL sequence. FLOPs denote total MACs.
Inference latency is per image (batch\,=\,1).
Best results highlighted for \textcolor{green!80}{buffer-free} and \textcolor{cyan!70}{memory-based}.}
\label{tab:compute_all}
\centering
\setlength{\tabcolsep}{3pt}
\resizebox{.95\textwidth}{!}{\begin{tabular}{@{}l|rrrrr|rrrrr|rrrrr@{}}
\toprule
& \multicolumn{5}{c|}{\textbf{CIFAR-100}}
& \multicolumn{5}{c|}{\textbf{Tiny-ImageNet}}
& \multicolumn{5}{c}{\textbf{ImageNet-1000}} \\
Method
& FLOPs$\downarrow$ & Train$\downarrow$ & GPU$\downarrow$ & Infer$\downarrow$ & ACC$\uparrow$
& FLOPs$\downarrow$ & Train$\downarrow$ & GPU$\downarrow$ & Infer$\downarrow$ & ACC$\uparrow$
& FLOPs$\downarrow$ & Train$\downarrow$ & GPU$\downarrow$ & Infer$\downarrow$ & ACC$\uparrow$ \\
& (TF) & Time & (GB) & (ms) & (\%)
& (TF) & Time & (GB) & (ms) & (\%)
& (PF) & Time & (GB) & (ms) & (\%) \\
\midrule
\multicolumn{16}{c}{\textit{Memory-free methods}} \\
\midrule
SI
& \cellcolor{green!20}\textbf{4,270} & \cellcolor{green!20}\textbf{24 min} & \cellcolor{green!20}\textbf{2.0} & \cellcolor{green!20}\textbf{0.4} & 15.37
& \cellcolor{green!20}\textbf{8,541} & \cellcolor{green!20}\textbf{1.0 h} & \cellcolor{green!20}\textbf{2.4} & \cellcolor{green!20}\textbf{0.6} & 13.37
& \cellcolor{green!20}\textbf{716} & \cellcolor{green!20}\textbf{18 h} & \cellcolor{green!20}\textbf{4.4} & \cellcolor{green!20}\textbf{2.5} & 9.68 \\
EWC
& 4,300 & 25 min & 2.3 & 0.4 & 12.87
& 8,610 & 1.1 h & 2.7 & 0.6 & 10.87
& 718 & 19 h & 4.8 & 2.5 & 7.53 \\
LwF
& 5,550 & 31 min & 2.2 & 0.4 & 19.56
& 11,100 & 1.3 h & 2.6 & 0.6 & 15.56
& 930 & 23 h & 4.6 & 2.5 & 11.24 \\
SpaceNet
& 5,650 & 33 min & 2.3 & 0.4 & 27.10
& 11,300 & 1.4 h & 2.7 & 0.6 & 17.80
& 947 & 25 h & 4.7 & 2.5 & 13.50 \\
PEC
& 5,374 & 45 min & 2.5 & 3.2 & 29.40
& 10,748 & 1.9 h & 2.9 & 5.8 & 19.40
& 901 & 34 h & 4.9 & 18.5 & 14.83 \\
NFL
& 6,735 & 37 min & 2.4 & 0.4 & 41.20
& 13,469 & 1.6 h & 2.8 & 0.6 & 32.50
& 1,129 & 28 h & 4.8 & 2.5 & 27.15 \\
NISPA
& 9,100 & 1.0 h & 3.0 & 0.5 & 43.80
& 18,200 & 2.6 h & 3.4 & 0.7 & 35.40
& 1,526 & 48 h & 5.4 & 2.8 & 29.40 \\
DCNet
& 6,850 & 40 min & 2.5 & 0.5 & 52.85
& 13,700 & 1.7 h & 2.9 & 0.7 & 43.90
& 1,148 & 30 h & 4.9 & 2.6 & 37.80 \\
\textbf{NFL+}
& 7,286 & 50 min & 2.6 & 0.5 & \cellcolor{green!20}\textbf{53.70}
& 14,573 & 2.1 h & 3.0 & 0.7 & \cellcolor{green!20}\textbf{44.70}
& 1,221 & 38 h & 5.0 & 2.7 & \cellcolor{green!20}\textbf{38.42} \\
\midrule
\multicolumn{16}{c}{\textit{Memory-based methods}} \\
\midrule
DER++
& \cellcolor{cyan!20}\textbf{6,504} & \cellcolor{cyan!20}\textbf{28 min} & 3.6 & \cellcolor{cyan!20}\textbf{0.4} & 21.20
& \cellcolor{cyan!20}\textbf{10,590} & \cellcolor{cyan!20}\textbf{1.2 h} & 4.0 & \cellcolor{cyan!20}\textbf{0.6} & 18.10
& \cellcolor{cyan!20}\textbf{843} & \cellcolor{cyan!20}\textbf{21 h} & 6.0 & \cellcolor{cyan!20}\textbf{2.5} & 14.20 \\
iCaRL
& 6,188 & 1.4 h & \cellcolor{cyan!20}\textbf{3.3} & 0.8 & 21.50
& 10,275 & 3.5 h & \cellcolor{cyan!20}\textbf{3.7} & 1.2 & 15.80
& 823 & 64 h & \cellcolor{cyan!20}\textbf{5.7} & 4.5 & 12.45 \\
MEMO
& 14,340 & 51 min & 6.7 & 1.4 & 54.40
& 23,688 & 2.2 h & 7.0 & 2.3 & 46.20
& 1,894 & 40 h & 10.6 & 8.5 & 38.90 \\
DyTox
& 25,092 & 1.7 h & 5.8 & 1.9 & \cellcolor{cyan!20}\textbf{55.80}
& 41,356 & 4.7 h & 7.2 & 3.5 & \cellcolor{cyan!20}\textbf{47.60}
& 3,304 & 95 h & 14.3 & 11.2 & \cellcolor{cyan!20}\textbf{40.15} \\
\bottomrule
\end{tabular}
}
\vspace{0.3em}
{\footnotesize TF=TFLOPs. PF=PFLOPs.
Infer\,(ms): per-image latency at batch\,=\,1 on A6000.
DyTox/MEMO FLOPs include growing model overhead.}
\end{table*}
\begin{table}[t]
\caption{Efficiency analysis of NFL+ vs DyTox and MEMO.}
\label{tab:efficiency}
\centering
\small
\setlength{\tabcolsep}{4pt}
\begin{tabular}{@{}lccc@{}}
\toprule
 & CIFAR & Tiny- & ImageNet \\
 & -100  & ImgNet & -1000 \\
\midrule
\multicolumn{4}{c}{\textit{NFL+ vs.\ DyTox}} \\
\midrule
ACC gap              & 2.10\%       & 2.90\%       & 1.73\% \\
FLOPs ratio          & 0.29$\times$ & 0.35$\times$ & 0.37$\times$ \\
Training speedup     & 2.0$\times$  & 2.2$\times$  & 2.5$\times$ \\
Inference speedup    & 3.8$\times$  & 5.0$\times$  & 4.1$\times$ \\
Peak GPU mem saved   & 8.0 GB       & 7.2 GB       & 9.3 GB \\
\midrule
\multicolumn{4}{c}{\textit{NFL+ vs.\ MEMO}} \\
\midrule
ACC gap              & 0.70\%       & 1.50\%       & 0.48\% \\
FLOPs ratio          & 0.51$\times$ & 0.62$\times$ & 0.64$\times$ \\
Training speedup     & 1.0$\times$  & 1.0$\times$  & 1.1$\times$ \\
Inference speedup    & 2.8$\times$  & 3.3$\times$  & 3.1$\times$ \\
Peak GPU mem saved   & 4.1 GB       & 4.0 GB       & 5.6 GB \\
\bottomrule
\end{tabular}
\end{table}
Table~\ref{tab:efficiency} quantifies the efficiency--accuracy trade-off of NFL+ against the two strongest memory-based competitors. Relative to DyTox, NFL+ closes the accuracy gap to 1.73--2.90\% across all three benchmarks while requiring only 0.29--0.37$\times$ the FLOPs, training 2.0--2.5$\times$ faster, running inference 3.8--5.0$\times$ faster, and consuming 7.2--9.3\,GB less peak GPU memory. The comparison with MEMO is particularly striking: on ImageNet-1000, the accuracy gap narrows to just 0.48\%, yet NFL+ uses 0.64$\times$ the FLOPs, achieves 3.1$\times$ lower inference latency, and saves 5.6\,GB of GPU memory. Crucially, both DyTox and MEMO additionally require storing an exemplar buffer throughout the CL sequence (4,797\,MB and 3,647\,MB respectively), a cost that NFL+ eliminates entirely by design. These results demonstrate that the multi-step freezing strategy of NFL+ achieves a fundamentally more favorable compute--accuracy Pareto frontier than replay-based alternatives: it trades a marginal accuracy deficit for substantial reductions in computation, memory, and storage, while simultaneously removing the privacy and scalability constraints imposed by exemplar retention.

\subsection{Parameter Usage Efficiency Analysis}\label{Parameter Usage Efficiency}
\begin{figure*}[t]
    \centering
    \includegraphics[width=\linewidth]{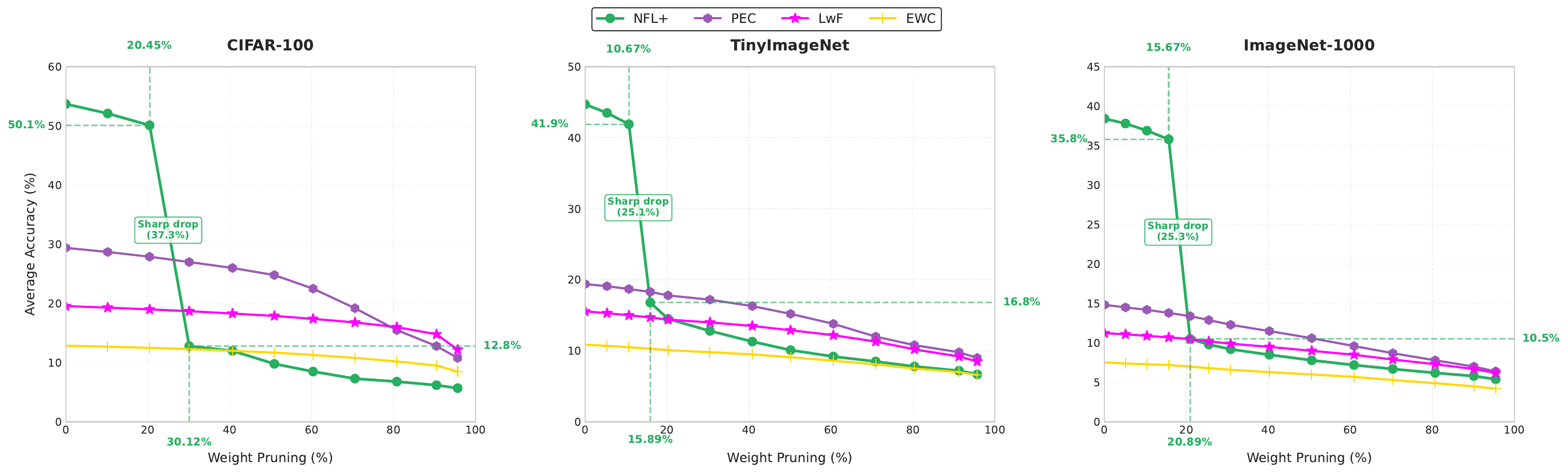}
    \caption{Parameter usage efficiency across three datasets under CIL with 10 tasks. The dashed vertical lines indicate the critical pruning percentage where each method experiences significant accuracy degradation. Sharper declines at lower pruning levels imply more efficient use of essential network parameters.}
\label{fig:weight_pruning}
\end{figure*}
We analyze weight utilization patterns across CL methods through progressive weight pruning, as illustrated in Fig.~\ref{fig:weight_pruning}. NFL+ exhibits a distinctive sharp accuracy drop at specific pruning thresholds: on CIFAR-100, accuracy plummets from 50.1\% to 12.8\% (37.3\% drop) when pruning increases from 20.45\% to 30.12\%; on Tiny-ImageNet, from 41.9\% to 16.8\% (25.1\% drop) between 10.67\% and 15.89\% pruning; and on ImageNet-1000, from 35.8\% to 10.5\% (25.3\% drop) between 15.67\% and 20.89\% pruning. In contrast, PEC exhibits monotonic accuracy decline across the entire pruning range. On CIFAR-100, accuracy decreases smoothly from 29.40\% to 10.8\% over 0--95\% pruning, with no identifiable critical threshold. This robustness to pruning exposes PEC's fundamental inefficiency: training $|\mathcal{N}|$ separate student networks without exploiting class similarity creates massive parameter redundancy. 

Nevertheless, LwF and EWC exhibit gradual degradation across all pruning levels. On CIFAR-100, LwF declines smoothly from 19.56\% to 12.2\% across the full pruning range, while EWC drops from 12.87\% to 8.5\%, neither showing abrupt transitions.
These findings suggest that NFL+'s stepwise freezing strategy effectively isolates task knowledge in specific weight subsets, making it inherently sparse-friendly while using fewer parameters more effectively than approaches that wastefully allocate redundant capacity.

\subsection{Ablation Study of the NFL and the NFL+}\label{Ablation Study of the NFL and the NFL+}

\begin{table}[t]
\centering
\caption{Ablation study of each NFL step on CIFAR-100 (CIL scenario with 10 tasks). Each row adds the corresponding step to the pipeline.}
\label{tab:NFL_ablation}
\vspace{0.5em}
\small
\setlength{\tabcolsep}{6pt}
\renewcommand{\arraystretch}{1.15}
\begin{tabular}{@{}lcc@{}}
\toprule
\textbf{Step Configuration} & $\Delta$ (\%) & ACC (\%) \\
\midrule
Step 1: Basic training on $T_t$                      & --                        & 10.12 \\
Step 2: Task-specific head training                  & +3.85                     & 13.97 \\
Step 3: Knowledge distillation                       & +4.28                     & 18.25 \\
Step 4: Fine-tuning with dual logit targets          & \cellcolor{bestcell}+22.95 & \cellcolor{bestcell}41.20 \\
\bottomrule
\end{tabular}
\end{table}

\begin{table}[t]
\centering
\caption{Ablation study of each NFL+ step on CIFAR-100 (CIL scenario with 10 tasks). Each row adds the corresponding step to the pipeline. Step~2 trains the auxiliary Auto-Encoder on frozen representations; since no classification parameters are updated, accuracy is unchanged.}
\label{tab:NFL_plus_ablation}
\vspace{0.5em}
\small
\setlength{\tabcolsep}{6pt}
\renewcommand{\arraystretch}{1.15}
\begin{tabular}{@{}lcc@{}}
\toprule
\textbf{Step Configuration} & $\Delta$ (\%) & ACC (\%) \\
\midrule
Step 1: Basic training on $T_t$                       & --                        & 10.12 \\
Step 2: Auto-Encoder training                         & 0.00                      & 10.12 \\
Step 3: Task-specific head training                   & +3.87                     & 13.99 \\
Step 4: Knowledge distillation                        & +8.45                     & 22.44 \\
Step 5: Fine-tuning with bias correction              & \cellcolor{bestcell}+31.26 & \cellcolor{bestcell}53.70 \\
\bottomrule
\end{tabular}
\end{table}

\begin{table}[t]
\centering
\caption{Component-removal ablation for NFL+ on CIFAR-100 (CIL, 10 tasks). Each row removes one component from the full NFL+ pipeline.}
\label{tab:NFL_plus_component}
\vspace{0.5em}
\small
\setlength{\tabcolsep}{6pt}
\renewcommand{\arraystretch}{1.15}
\begin{tabular}{@{}lcc@{}}
\toprule
\textbf{Configuration} & ACC (\%) & Drop (\%) \\
\midrule
Full NFL+ (Steps 1--5)                               & \cellcolor{bestcell}\textbf{53.70} & -- \\
\;\;w/o bias correction ($\widetilde{H}_t \to H_t$)  & 47.85 & $-$5.85 \\
\;\;w/o AE feature constraint                        & 49.20 & $-$4.50 \\
\;\;w/o both AE components (= NFL Step 4)             & 41.20 & $-$12.50 \\
\;\;w/o new head isolation (skip Step 3)              & 50.10 & $-$3.60 \\
\bottomrule
\end{tabular}
\end{table}

Tables~\ref{tab:NFL_ablation} and~\ref{tab:NFL_plus_ablation} present sequential ablation studies for NFL and NFL+, illustrating the contribution of each pipeline step to the final performance.

\textbf{NFL.} As shown in Table~\ref{tab:NFL_ablation}, the baseline model, corresponding to basic training on the current task~$T_t$, achieves an accuracy of 10.12\%. Steps~2 and~3 yield incremental improvements: task-specific head training increases performance by +3.85\%, while knowledge distillation contributes an additional +4.28\%, resulting in a cumulative accuracy of 18.25\%. The most significant gain arises from Step~4, fine-tuning with dual logit targets, which delivers a +22.95\% improvement, nearly three times the combined contribution of the preceding steps. This result highlights dual-logit fine-tuning as the core mechanism of NFL, effectively consolidating representations learned in earlier stages.

\textbf{NFL+.} Table~\ref{tab:NFL_plus_ablation} reports the sequential ablation for NFL+. Step~2 trains the auxiliary Auto-Encoder on frozen backbone representations; since neither the backbone nor the classifier is updated, classification accuracy remains at 10.12\%. The Auto-Encoder's contribution materializes in Step~5, where it provides both the feature-space constraint (anchoring encoded representations to their pre-adaptation values) and the bias correction mechanism. Steps~3 and~4 contribute +3.87\% and +8.45\%, bringing accuracy to 22.44\%. Step~5 delivers the largest single-step gain of +31.26\%, reflecting the combined effect of bias-corrected distillation, the AE feature constraint, and the new-task supervision acting on the foundation laid by all preceding steps.

To disentangle the contributions within Step~5, Table~\ref{tab:NFL_plus_component} presents a component-removal study. Removing bias correction alone (replacing $\widetilde{H}_t$ with the uncorrected logits $H_t$) reduces accuracy by 5.85 points, confirming that explicitly addressing task-recency bias is critical in the CIL setting. Removing the AE feature constraint alone costs 4.50 points, indicating that anchoring representations to the pre-adaptation manifold provides complementary regularization. Removing both AE components reduces NFL+ to the NFL Step~4 loss structure, recovering exactly NFL's accuracy of 41.20\%. Notably, the combined drop ($-$12.50\%) exceeds the sum of individual drops ($-$10.35\%), revealing a synergy: the feature constraint preserves the representation geometry on which the bias correction operates, and the bias correction enables the distillation signal to focus on discrimination rather than compensating for class imbalance. Skipping head isolation (Step~3) costs 3.60 points, confirming that initializing the new head on frozen features before backbone adaptation remains beneficial even with the AE components present.

\section{Pseudo-code of NFL, NFL+ and NFL+LoRA}\label{Pseudo-code of the NFL and the NFL+}
In this section, we present the pseudo-code for the NFL and the NFL+ to clarify the shared steps and highlight their key differences. As discussed in the main body of the paper, the NFL+ integrates an Auto-Encoder after the learning task $T_t$ to preserve the most informative features learned thus far. Overall, Steps 3 and 4 in NFL+, Algorithm~\ref{alg: NFL+}, correspond to Steps 2 and 3 in the NFL, Algorithm~\ref{alg: NFL}.

\begin{algorithm}[t]
\caption{\textbf{NFL: Base Stepwise Freezing}}\label{alg: NFL}
\begin{algorithmic}[1]
\REQUIRE Trained model $\text{NN}^1(\theta^*_s, \theta^*_t)$ on task $T_t$; new task data $\mathcal{D}_{t+1} = (X_{t+1}, Y_{t+1})$
\ENSURE Model for $T_t \cup T_{t+1}$

\STATE \textbf{Step 1: Store soft targets}
\STATE $H_t \leftarrow \text{NN}^1(X_{t+1};\, \theta^*_s, \theta^*_t)$ 

\STATE \textbf{Step 2: New head initialization} \hfill \textit{Config: FFT}
\STATE Freeze $\theta^*_s,\, \theta^*_t$; train $\theta_{t+1}$ via $\mathcal{L}^2 = \mathcal{L}_{\text{CE}}(Y_{t+1},\, O^2_{t+1})$

\STATE \textbf{Step 3: Controlled backbone adaptation} \hfill \textit{Config: TTF}
\STATE Freeze $\theta^*_{t+1}$; train $\theta_s,\, \theta_t$ via $\mathcal{L}^3 = \mathcal{L}_{\text{KD}}(H_t,\, O^3_t) + \mathcal{L}_{\text{CE}}(Y_{t+1},\, O^3_{t+1})$

\STATE \textbf{Step 4: Joint fine-tuning} \hfill \textit{Config: TTT}
\STATE $H^\prime_t \leftarrow \text{NN}(X_{t+1};\, \theta^u_s, \theta_t)$ 
\STATE Unfreeze all parameters; train via:
\STATE \quad $\mathcal{L}^4 = \alpha\,\mathcal{L}_{\text{KD}}(H_t,\, O^4_t) + (1 - \alpha)\,\mathcal{L}_{\text{KD}}(H^\prime_t,\, O^{\prime 4}_t) + \mathcal{L}_{\text{CE}}(Y_{t+1},\, O^4_{t+1})$

\STATE \textbf{return} $\text{NN}^4(\theta_s, \theta_t, \theta_{t+1})$
\end{algorithmic}
\end{algorithm}

\begin{algorithm}[t]
\caption{\textbf{NFL+: Stepwise Freezing with Auto-Encoder Regularization}}\label{alg: NFL+}
\begin{algorithmic}[1]
\REQUIRE Trained model $\text{NN}^1(\theta^*_s, \theta^*_t)$ on task $T_t$;
Auto-Encoder $R$ trained at the end of $T_t$;
new task data $\mathcal{D}_{t+1} = (X_{t+1}, Y_{t+1})$
\ENSURE Model for $T_t \cup T_{t+1}$; updated Auto-Encoder for $T_{t+1}$

\STATE \textbf{Step 1: Store soft targets}
\STATE $H_t \leftarrow \text{NN}^1(X_{t+1};\, \theta^*_s, \theta^*_t)$

\STATE \textbf{Step 2: New head initialization} \hfill \textit{Config: FFT}
\STATE Freeze $\theta^*_s,\, \theta^*_t$; train $\theta_{t+1}$ via $\mathcal{L}^2 = \mathcal{L}_{\text{CE}}(Y_{t+1},\, O^2_{t+1})$

\STATE \textbf{Step 3: Controlled backbone adaptation} \hfill \textit{Config: TTF}
\STATE Freeze $\theta^*_{t+1}$; train $\theta_s,\, \theta_t$ via $\mathcal{L}^3 = \mathcal{L}_{\text{KD}}(H_t,\, O^3_t) + \mathcal{L}_{\text{CE}}(Y_{t+1},\, O^3_{t+1})$

\STATE \textbf{Step 4: Bias-corrected joint fine-tuning} \hfill \textit{Config: TTT}
\STATE Train bias layer on held-out set: $\widetilde{H}_t = \Gamma(P_{t+1}) \odot H_t$, \quad regularized by $\|\Gamma(P_{t+1}) - \mathbf{1}\|^2_2$
\STATE $H^\prime_t \leftarrow \text{NN}(X_{t+1};\, \theta^u_s, \theta_t)$
\STATE Unfreeze all parameters; train via:
\STATE \quad $\mathcal{L}^4 = \eta\,\mathcal{L}_{\text{KD}}(\widetilde{H}_t,\, O^4_t) + (1 - \eta)\,\mathcal{L}_{\text{KD}}(H^\prime_t,\, O^{\prime 4}_t) + \|\sigma(W_{\text{Enc}}\,\theta_s) - \sigma(W_{\text{Enc}}\,\theta^*_s)\|^2_2 + \mathcal{L}_{\text{CE}}(Y_{t+1},\, O^4_{t+1})$

\STATE \textbf{Step 5: Prepare Auto-Encoder for next task}
\STATE $P_{t+1} \leftarrow \theta_s(X_{t+1})$
\STATE Retrain $R$ via $\min_{R}\; \Omega\,\|R(P_{t+1}) - P_{t+1}\|^2_2 + \mathcal{L}_{\text{CE}}\big(\theta_{t+1}(R(P_{t+1})),\, Y_{t+1}\big)$

\STATE \textbf{return} $\text{NN}^4(\theta_s, \theta_t, \theta_{t+1}, R, \Gamma)$
\end{algorithmic}
\end{algorithm}

\begin{algorithm}[t]
\caption{\textbf{NFL+LoRA: Stepwise Freezing for Pre-trained Vision Transformers}}\label{alg: NFL+LoRA}
\begin{algorithmic}[1]
\REQUIRE Accumulated base weights $\mathbf{W}_{t-1}$; LoRA rank $r$; cumulative Fisher $\mathbf{F}^{\text{cum}}_{t-1}$; decay $\gamma$
\REQUIRE Trained LoRA parameters $\mathbf{A}^*_{t-1}, \mathbf{B}^*_{t-1}$ and head $\theta^*_t$ from task $T_t$; new task data $\mathcal{D}_{t+1} = (X_{t+1}, Y_{t+1})$
\ENSURE Model for $T_t \cup T_{t+1}$

\STATE \textbf{Step 1: Store soft targets}
\STATE $H_t \leftarrow \text{NN}^1(X_{t+1};\, \mathbf{W}_{t-1} + \mathbf{A}^*_{t-1}\mathbf{B}^*_{t-1},\, \theta^*_t)$ 
\STATE \textbf{Step 2: Merge, reset, and initialize new head} \hfill \textit{Config: FFT}
\STATE $\mathbf{W}_t \leftarrow \mathbf{W}_{t-1} + \mathbf{A}^*_{t-1}\mathbf{B}^*_{t-1}$ 
\STATE $\mathbf{F}^{\text{cum}}_t \leftarrow \gamma\,\mathbf{F}^{\text{cum}}_{t-1} + \mathbf{F}_t$ 
\STATE Initialize fresh $\mathbf{A}, \mathbf{B} \leftarrow \mathbf{0}$
\STATE Freeze $\mathbf{W}_t, \mathbf{A}, \mathbf{B}, \theta^*_t$; train $\theta_{t+1}$ via $\mathcal{L}^2 = \mathcal{L}_{\text{CE}}(Y_{t+1},\, O^2_{t+1})$

\STATE \textbf{Step 3: LoRA adaptation} \hfill \textit{Config: TTF}
\STATE Freeze $\theta^*_{t+1}$; train $\mathbf{A}, \mathbf{B}, \theta_t$ via $\mathcal{L}^3 = \mathcal{L}_{\text{KD}}(H_t,\, O^3_t) + \mathcal{L}_{\text{CE}}(Y_{t+1},\, O^3_{t+1})$

\STATE \textbf{Step 4: Joint fine-tuning with Fisher regularization} \hfill \textit{Config: TTT}
\STATE $H^\prime_t \leftarrow \text{NN}(X_{t+1};\, \mathbf{W}_t + \mathbf{A}^u\mathbf{B}^u,\, \theta_t)$ 
\STATE Unfreeze $\mathbf{A}, \mathbf{B}$ and all heads; train via:
\STATE \quad $\mathcal{L}^4 = \alpha\,\mathcal{L}_{\text{KD}}(H_t,\, O^4_t) + (1 - \alpha)\,\mathcal{L}_{\text{KD}}(H^\prime_t,\, O^{\prime 4}_t) + \mathcal{L}_{\text{CE}}(Y_{t+1},\, O^4_{t+1}) + \frac{\lambda}{2}\,\text{vec}(\mathbf{A}\mathbf{B})^\top \mathbf{F}^{\text{cum}}_t\,\text{vec}(\mathbf{A}\mathbf{B})$

\STATE \textbf{return} $\text{NN}^4(\mathbf{W}_t, \mathbf{A}^*, \mathbf{B}^*, \theta_t, \theta_{t+1})$
\end{algorithmic}
\end{algorithm}
\section{Implementation Details of iCaRL}\label{Implementation Details of iCaRL}

This section outlines our adaptation of iCaRL~\citep{rebuffi2017icarl} for the TIL setting, which builds upon its original proposal for CIL. A key modification involves refining its classification mechanism.

Conventionally, iCaRL determines a label $y^*$ by identifying the class whose average exemplar feature vector is most proximate to the input example's feature vector. Specifically, given $\overline{\mathbf{y}}$ as the average feature vector of exemplars for class $y$ and ${\phi}(\mathbf{x})$ as the feature vector derived from example $\mathbf{x}$, iCaRL's prediction is formulated as:
\begin{equation}
y^* = \underset{y=1,\dots,t}{\operatorname{argmin}} \|{\phi}(\mathbf{x}) - \overline{\mathbf{y}}\|_2.
\label{eq:icarl_original_academic}
\end{equation}
Our modified approach, however, casts iCaRL's network response, $h(\mathbf{x})$, in terms of the negative Euclidean distance to the tensor of average feature vectors for all classes, ${\mathbf{\Phi}}$. This yields:
\begin{equation}
h(\mathbf{x}) = -\|{\phi}(\mathbf{x}) - {\mathbf{\Phi}}\|_2.
\label{eq:icarl_modified_response_academic}
\end{equation}
It is pertinent to note that when considering the argmax of $h(\mathbf{x})$ in a CIL context, this formulation yields an identical prediction to that of Eq.~\ref{eq:icarl_original_academic}. Furthermore, it is noteworthy that iCaRL incorporates a weight-decay regularization term, a crucial element in ensuring its competitive performance compared with other proposed approaches.

\section{Hyperparameter Search}\label{Hyperparameter Search}
Hyperparameters are selected using only the first task's validation split (10\% of the first task's data) following the \citep{cha2025hyperparameters}. All baselines are tuned under the same first-task protocol to ensure a fair comparison. Results are averaged over 10 runs with different task orderings to account for ordering sensitivity, as recommended by~\citet{lee2024hyperparameter}. We show the complete hyperparameter space in Table~\ref{tab:hyperparameters_complete} and the best hyperparameter combination that we chose for each method for the experiments in the main paper, Table~\ref{tab:best_hyperparameters}. We denote the learning rate with $lr$, weight decay with $wd$, and Adam optimizer hyperparameters for Auto-Encoder training with $lr_{adam},\epsilon,\beta_1$, and $\beta_2$.
\begin{table*}[t]
\centering
\caption{Complete hyperparameter search space for all methods across datasets. All methods use the Adam optimizer with shared parameters $\beta_1 \in \{0.9\}$, $\beta_2 \in \{0.999\}$, $\epsilon \in \{10^{-8}\}$. For CIL, memory-based methods use a buffer of 2,000 (CIFAR-100, Tiny-ImageNet) or 20,000 (ImageNet-1000). For TIL, the buffer is fixed at 200.}
\label{tab:hyperparameters_complete}
\setlength{\tabcolsep}{2.5pt}
\tiny
\begin{tabular}{llll}
\toprule
\textbf{Method} & \textbf{Scenario} & \textbf{Dataset} & \textbf{Hyperparameters (searched on first task only)} \\
\midrule
\multicolumn{4}{c}{\textit{\textbf{ResNet-18 backbone}}} \\
\midrule
\multirow{3}{*}{NFL} & \multirow{3}{*}{CIL/TIL} & CIFAR-100 & $lr$: [0.0001, 0.001, 0.01, 0.03, 0.1], $\quad p$: [2.0, 3.0, 4.0], $\quad \alpha$: [0.3, 0.5, 0.7] \\
 &  & Tiny-ImageNet & $lr$: [0.0001, 0.001, 0.01, 0.03, 0.1], $\quad p$: [2.0, 3.0, 4.0], $\quad \alpha$: [0.3, 0.5, 0.7] \\
 &  & ImageNet-1000 & $lr$: [0.0001, 0.001, 0.01, 0.03, 0.1], $\quad p$: [2.0, 3.0, 4.0], $\quad \alpha$: [0.3, 0.5, 0.7] \\
\midrule
\multirow{6}{*}{NFL+} & \multirow{6}{*}{CIL/TIL} & CIFAR-100 & $lr$: [0.0001, 0.001, 0.01, 0.03, 0.1], $\quad p$: [2.0, 3.0, 4.0], $\quad \Omega$: [0.1, 0.3, 0.5, 1.0], $\quad \eta$: [0.3, 0.5, 0.7] \\
 &  &  & $lr_{\text{ae}}$: [0.0001, 0.001, 0.01] \\
 &  & Tiny-ImageNet & $lr$: [0.0001, 0.001, 0.01, 0.03, 0.1], $\quad p$: [2.0, 3.0, 4.0], $\quad \Omega$: [0.1, 0.3, 0.5, 1.0], $\quad \eta$: [0.3, 0.5, 0.7] \\
 &  &  & $lr_{\text{ae}}$: [0.0001, 0.001, 0.01] \\
 &  & ImageNet-1000 & $lr$: [0.0001, 0.001, 0.01, 0.03, 0.1], $\quad p$: [2.0, 3.0, 4.0], $\quad \Omega$: [0.1, 0.3, 0.5, 1.0], $\quad \eta$: [0.3, 0.5, 0.7] \\
 &  &  & $lr_{\text{ae}}$: [0.0001, 0.001, 0.01] \\
\midrule
\multirow{3}{*}{DCNet} & \multirow{3}{*}{CIL/TIL} & CIFAR-100 & $lr$: [0.0001, 0.001, 0.01, 0.03, 0.1], $\quad \lambda_{\text{dc}}$: [0.1, 0.5, 1.0, 5.0] \\
 &  & Tiny-ImageNet & $lr$: [0.0001, 0.001, 0.01, 0.03, 0.1], $\quad \lambda_{\text{dc}}$: [0.1, 0.5, 1.0, 5.0] \\
 &  & ImageNet-1000 & $lr$: [0.0001, 0.001, 0.01, 0.03, 0.1], $\quad \lambda_{\text{dc}}$: [0.1, 0.5, 1.0, 5.0] \\
\midrule
\multirow{3}{*}{NISPA} & \multirow{3}{*}{CIL/TIL} & CIFAR-100 & $lr$: [0.0001, 0.001, 0.01, 0.03, 0.1], $\quad s$: [0.3, 0.5, 0.7, 0.9], $\quad \lambda_{\text{reg}}$: [0.1, 1.0, 10.0] \\
 &  & Tiny-ImageNet & $lr$: [0.0001, 0.001, 0.01, 0.03, 0.1], $\quad s$: [0.3, 0.5, 0.7, 0.9], $\quad \lambda_{\text{reg}}$: [0.1, 1.0, 10.0] \\
 &  & ImageNet-1000 & $lr$: [0.0001, 0.001, 0.01, 0.03, 0.1], $\quad s$: [0.3, 0.5, 0.7, 0.9], $\quad \lambda_{\text{reg}}$: [0.1, 1.0, 10.0] \\
\midrule
\multirow{3}{*}{WSN} & \multirow{3}{*}{TIL} & CIFAR-100 & $lr$: [0.0001, 0.001, 0.01, 0.03] \\
 &  & Tiny-ImageNet & $lr$: [0.0001, 0.001, 0.01, 0.03] \\
 &  & ImageNet-1000 & $lr$: [0.0001, 0.001, 0.01, 0.03] \\
\midrule
\multirow{3}{*}{PEC} & \multirow{3}{*}{CIL/TIL} & CIFAR-100 & $lr$: [0.0001, 0.001, 0.01, 0.03], $\quad \lambda_{\text{pec}}$: [0.1, 0.5, 1.0, 5.0] \\
 &  & Tiny-ImageNet & $lr$: [0.0001, 0.001, 0.01, 0.03], $\quad \lambda_{\text{pec}}$: [0.1, 0.5, 1.0, 5.0] \\
 &  & ImageNet-1000 & $lr$: [0.0001, 0.001, 0.01, 0.03], $\quad \lambda_{\text{pec}}$: [0.1, 0.5, 1.0, 5.0] \\
\midrule
\multirow{3}{*}{SpaceNet} & \multirow{3}{*}{CIL/TIL} & CIFAR-100 & $lr$: [0.0001, 0.001, 0.01, 0.03], $\quad s_{\text{init}}$: [0.3, 0.5, 0.7], $\quad \lambda_{\text{sp}}$: [0.1, 1.0, 10.0] \\
 &  & Tiny-ImageNet & $lr$: [0.0001, 0.001, 0.01, 0.03], $\quad s_{\text{init}}$: [0.3, 0.5, 0.7], $\quad \lambda_{\text{sp}}$: [0.1, 1.0, 10.0] \\
 &  & ImageNet-1000 & $lr$: [0.0001, 0.001, 0.01, 0.03], $\quad s_{\text{init}}$: [0.3, 0.5, 0.7], $\quad \lambda_{\text{sp}}$: [0.1, 1.0, 10.0] \\
\midrule
\multirow{3}{*}{LwF} & \multirow{3}{*}{CIL/TIL} & CIFAR-100 & $lr$: [0.0001, 0.001, 0.01, 0.03, 0.1], $\quad \alpha$: [0.3, 0.5, 1.0, 3.0], $\quad T$: [2.0, 4.0], $\quad wd$: [1e-5, 5e-5] \\
 &  & Tiny-ImageNet & $lr$: [0.0001, 0.001, 0.01, 0.03, 0.1], $\quad \alpha$: [0.3, 0.5, 1.0, 3.0], $\quad T$: [2.0, 4.0], $\quad wd$: [1e-5, 5e-5] \\
 &  & ImageNet-1000 & $lr$: [0.0001, 0.001, 0.01, 0.03, 0.1], $\quad \alpha$: [0.3, 0.5, 1.0, 3.0], $\quad T$: [2.0, 4.0], $\quad wd$: [1e-5, 5e-5] \\
\midrule
\multirow{3}{*}{EWC} & \multirow{3}{*}{CIL/TIL} & CIFAR-100 & $lr$: [0.0001, 0.001, 0.01, 0.03, 0.1], $\quad \lambda$: [10, 25, 30, 90, 100], $\quad \gamma$: [0.9, 0.95, 1.0] \\
 &  & Tiny-ImageNet & $lr$: [0.0001, 0.001, 0.01, 0.03, 0.1], $\quad \lambda$: [10, 25, 30, 90, 100], $\quad \gamma$: [0.9, 0.95, 1.0] \\
 &  & ImageNet-1000 & $lr$: [0.0001, 0.001, 0.01, 0.03, 0.1], $\quad \lambda$: [10, 25, 30, 90, 100], $\quad \gamma$: [0.9, 0.95, 1.0] \\
\midrule
\multirow{3}{*}{SI} & \multirow{3}{*}{CIL/TIL} & CIFAR-100 & $lr$: [0.0001, 0.001, 0.01, 0.03, 0.1], $\quad c$: [0.3, 0.5, 0.7, 1.0], $\quad \xi$: [0.9, 1.0] \\
 &  & Tiny-ImageNet & $lr$: [0.0001, 0.001, 0.01, 0.03, 0.1], $\quad c$: [0.3, 0.5, 0.7, 1.0], $\quad \xi$: [0.9, 1.0] \\
 &  & ImageNet-1000 & $lr$: [0.0001, 0.001, 0.01, 0.03, 0.1], $\quad c$: [0.3, 0.5, 0.7, 1.0], $\quad \xi$: [0.9, 1.0] \\
\midrule
\multirow{3}{*}{DyTox} & \multirow{3}{*}{CIL/TIL} & CIFAR-100 & $lr$: [0.0001, 0.001, 0.01, 0.03, 0.1] \\
 &  & Tiny-ImageNet & $lr$: [0.0001, 0.001, 0.01, 0.03, 0.1] \\
 &  & ImageNet-1000 & $lr$: [0.0001, 0.001, 0.01, 0.03, 0.1] \\
\midrule
\multirow{3}{*}{iCaRL} & \multirow{3}{*}{CIL/TIL} & CIFAR-100 & $lr$: [0.0001, 0.001, 0.01, 0.03, 0.1], $\quad wd$: [0, 1e-5, 5e-5, 1e-4] \\
 &  & Tiny-ImageNet & $lr$: [0.0001, 0.001, 0.01, 0.03, 0.1], $\quad wd$: [0, 1e-5, 5e-5, 1e-4] \\
 &  & ImageNet-1000 & $lr$: [0.0001, 0.001, 0.01, 0.03, 0.1], $\quad wd$: [0, 1e-5, 5e-5, 1e-4] \\
\midrule
\multirow{3}{*}{DER++} & \multirow{3}{*}{CIL/TIL} & CIFAR-100 & $lr$: [0.0001, 0.001, 0.01, 0.03, 0.1], $\quad \alpha$: [0.1, 0.2, 0.3, 0.5, 1.0], $\quad \beta$: [0.5, 1.0] \\
 &  & Tiny-ImageNet & $lr$: [0.0001, 0.001, 0.01, 0.03, 0.1], $\quad \alpha$: [0.1, 0.2, 0.3, 0.5, 1.0], $\quad \beta$: [0.5, 1.0] \\
 &  & ImageNet-1000 & $lr$: [0.0001, 0.001, 0.01, 0.03, 0.1], $\quad \alpha$: [0.1, 0.2, 0.3, 0.5, 1.0], $\quad \beta$: [0.5, 1.0] \\
\midrule
\multirow{3}{*}{MEMO} & \multirow{3}{*}{CIL/TIL} & CIFAR-100 & $lr$: [0.0001, 0.001, 0.01, 0.03, 0.1] \\
 &  & Tiny-ImageNet & $lr$: [0.0001, 0.001, 0.01, 0.03, 0.1] \\
 &  & ImageNet-1000 & $lr$: [0.0001, 0.001, 0.01, 0.03, 0.1] \\
\midrule
\multicolumn{4}{c}{\textit{\textbf{ViT-B/16 backbone (pretrained on ImageNet-21K)}}} \\
\midrule
\multirow{6}{*}{NFL+LoRA} & \multirow{6}{*}{CIL} & CIFAR-100 & $lr$: [1e-5, 5e-5, 1e-4, 5e-4, 1e-3], $\quad r$: [4, 8, 16], $\quad \lambda_{\text{F}}$: [0.1, 0.5, 1.0, 5.0], $\quad p$: [2.0, 3.0, 4.0] \\
 &  &  & $\alpha$: [0.3, 0.5, 0.7], $\quad \gamma$: [0.9, 0.95, 1.0] \\
 &  & ImageNet-R & $lr$: [1e-5, 5e-5, 1e-4, 5e-4, 1e-3], $\quad r$: [4, 8, 16], $\quad \lambda_{\text{F}}$: [0.1, 0.5, 1.0, 5.0], $\quad p$: [2.0, 3.0, 4.0] \\
 &  &  & $\alpha$: [0.3, 0.5, 0.7], $\quad \gamma$: [0.9, 0.95, 1.0] \\
 &  & ImageNet-A & $lr$: [1e-5, 5e-5, 1e-4, 5e-4, 1e-3], $\quad r$: [4, 8, 16], $\quad \lambda_{\text{F}}$: [0.1, 0.5, 1.0, 5.0], $\quad p$: [2.0, 3.0, 4.0] \\
 &  &  & $\alpha$: [0.3, 0.5, 0.7], $\quad \gamma$: [0.9, 0.95, 1.0] \\
\midrule
\multirow{3}{*}{CL-LoRA} & \multirow{3}{*}{CIL} & CIFAR-100 & $lr$: [1e-5, 5e-5, 1e-4, 5e-4, 1e-3], $\quad r$: [4, 8, 16] \\
 &  & ImageNet-R & $lr$: [1e-5, 5e-5, 1e-4, 5e-4, 1e-3], $\quad r$: [4, 8, 16] \\
 &  & ImageNet-A & $lr$: [1e-5, 5e-5, 1e-4, 5e-4, 1e-3], $\quad r$: [4, 8, 16] \\
\midrule
\multirow{3}{*}{EWC-LoRA} & \multirow{3}{*}{CIL} & CIFAR-100 & $lr$: [1e-5, 5e-5, 1e-4, 5e-4, 1e-3], $\quad r$: [4, 8, 16], $\quad \lambda$: [10, 25, 50, 100], $\quad \gamma$: [0.9, 0.95, 1.0] \\
 &  & ImageNet-R & $lr$: [1e-5, 5e-5, 1e-4, 5e-4, 1e-3], $\quad r$: [4, 8, 16], $\quad \lambda$: [10, 25, 50, 100], $\quad \gamma$: [0.9, 0.95, 1.0] \\
 &  & ImageNet-A & $lr$: [1e-5, 5e-5, 1e-4, 5e-4, 1e-3], $\quad r$: [4, 8, 16], $\quad \lambda$: [10, 25, 50, 100], $\quad \gamma$: [0.9, 0.95, 1.0] \\
\bottomrule
\end{tabular}
\end{table*}

\begin{table*}[t]
\centering
\caption{Selected best hyperparameters for all methods across datasets.}
\label{tab:best_hyperparameters}
\setlength{\tabcolsep}{3pt}
\small
\begin{tabular}{lllll}
\toprule
\textbf{Method} & \textbf{Scenario} & \textbf{Dataset} & \textbf{Selected Hyperparameters} \\
\midrule
\multicolumn{4}{c}{\textit{\textbf{ResNet-18 backbone}}} \\
\midrule
\multirow{3}{*}{NFL} & \multirow{3}{*}{CIL/TIL} & CIFAR-100 & $lr$: 0.01, $p$: 2.0, $\alpha$: 0.3 \\
 &  & Tiny-ImageNet & $lr$: 0.01, $p$: 2.0, $\alpha$: 0.3 \\
 &  & ImageNet-1000 & $lr$: 0.001, $p$: 2.0, $\alpha$: 0.5 \\
\midrule
\multirow{3}{*}{NFL+} & \multirow{3}{*}{CIL/TIL} & CIFAR-100 & $lr$: 0.01, $p$: 2.0, $\Omega$: 0.5, $\eta$: 0.5, $lr_{\text{ae}}$: 0.001 \\
 &  & Tiny-ImageNet & $lr$: 0.01, $p$: 2.0, $\Omega$: 0.5, $\eta$: 0.5, $lr_{\text{ae}}$: 0.001 \\
 &  & ImageNet-1000 & $lr$: 0.001, $p$: 2.0, $\Omega$: 0.5, $\eta$: 0.5, $lr_{\text{ae}}$: 0.001 \\
\midrule
\multirow{3}{*}{DCNet} & \multirow{3}{*}{CIL/TIL} & CIFAR-100 & $lr$: 0.01, $\lambda_{\text{dc}}$: 1.0 \\
 &  & Tiny-ImageNet & $lr$: 0.01, $\lambda_{\text{dc}}$: 1.0 \\
 &  & ImageNet-1000 & $lr$: 0.001, $\lambda_{\text{dc}}$: 0.5 \\
\midrule
\multirow{3}{*}{NISPA} & \multirow{3}{*}{CIL/TIL} & CIFAR-100 & $lr$: 0.01, $s$: 0.5, $\lambda_{\text{reg}}$: 1.0 \\
 &  & Tiny-ImageNet & $lr$: 0.01, $s$: 0.5, $\lambda_{\text{reg}}$: 1.0 \\
 &  & ImageNet-1000 & $lr$: 0.001, $s$: 0.5, $\lambda_{\text{reg}}$: 1.0 \\
\midrule
\multirow{3}{*}{WSN} & \multirow{3}{*}{TIL} & CIFAR-100 & $lr$: 0.001 \\
 &  & Tiny-ImageNet & $lr$: 0.001 \\
 &  & ImageNet-1000 & $lr$: 0.0001 \\
\midrule
\multirow{3}{*}{PEC} & \multirow{3}{*}{CIL} & CIFAR-100 & $lr$: 0.001, $\lambda_{\text{pec}}$: 1.0 \\
 &  & Tiny-ImageNet & $lr$: 0.001, $\lambda_{\text{pec}}$: 1.0 \\
 &  & ImageNet-1000 & $lr$: 0.0001, $\lambda_{\text{pec}}$: 0.5 \\
\midrule
\multirow{3}{*}{SpaceNet} & \multirow{3}{*}{CIL/TIL} & CIFAR-100 & $lr$: 0.001, $s_{\text{init}}$: 0.5, $\lambda_{\text{sp}}$: 1.0 \\
 &  & Tiny-ImageNet & $lr$: 0.001, $s_{\text{init}}$: 0.5, $\lambda_{\text{sp}}$: 1.0 \\
 &  & ImageNet-1000 & $lr$: 0.0001, $s_{\text{init}}$: 0.5, $\lambda_{\text{sp}}$: 1.0 \\
\midrule
\multirow{3}{*}{LwF} & \multirow{3}{*}{CIL/TIL} & CIFAR-100 & $lr$: 0.01, $\alpha$: 0.5, $T$: 2.0, $wd$: 5e-5 \\
 &  & Tiny-ImageNet & $lr$: 0.01, $\alpha$: 0.5, $T$: 2.0, $wd$: 5e-5 \\
 &  & ImageNet-1000 & $lr$: 0.001, $\alpha$: 0.5, $T$: 2.0, $wd$: 5e-5 \\
\midrule
\multirow{3}{*}{EWC} & \multirow{3}{*}{CIL/TIL} & CIFAR-100 & $lr$: 0.01, $\lambda$: 10, $\gamma$: 1.0 \\
 &  & Tiny-ImageNet & $lr$: 0.01, $\lambda$: 10, $\gamma$: 1.0 \\
 &  & ImageNet-1000 & $lr$: 0.001, $\lambda$: 25, $\gamma$: 1.0 \\
\midrule
\multirow{3}{*}{SI} & \multirow{3}{*}{CIL/TIL} & CIFAR-100 & $lr$: 0.01, $c$: 0.5, $\xi$: 1.0 \\
 &  & Tiny-ImageNet & $lr$: 0.01, $c$: 0.5, $\xi$: 1.0 \\
 &  & ImageNet-1000 & $lr$: 0.001, $c$: 0.5, $\xi$: 1.0 \\
\midrule
\multirow{3}{*}{DyTox} & \multirow{3}{*}{CIL/TIL} & CIFAR-100 & $lr$: 0.001 \\
 &  & Tiny-ImageNet & $lr$: 0.001 \\
 &  & ImageNet-1000 & $lr$: 0.0001 \\
\midrule
\multirow{3}{*}{iCaRL} & \multirow{3}{*}{CIL/TIL} & CIFAR-100 & $lr$: 0.01, $wd$: 5e-5 \\
 &  & Tiny-ImageNet & $lr$: 0.01, $wd$: 5e-5 \\
 &  & ImageNet-1000 & $lr$: 0.001, $wd$: 5e-5 \\
\midrule
\multirow{3}{*}{DER++} & \multirow{3}{*}{CIL/TIL} & CIFAR-100 & $lr$: 0.01, $\alpha$: 0.2, $\beta$: 0.5 \\
 &  & Tiny-ImageNet & $lr$: 0.01, $\alpha$: 0.2, $\beta$: 1.0 \\
 &  & ImageNet-1000 & $lr$: 0.001, $\alpha$: 0.2, $\beta$: 1.0 \\
\midrule
\multirow{3}{*}{MEMO} & \multirow{3}{*}{CIL/TIL} & CIFAR-100 & $lr$: 0.01 \\
 &  & Tiny-ImageNet & $lr$: 0.01 \\
 &  & ImageNet-1000 & $lr$: 0.001 \\
\midrule
\multicolumn{4}{c}{\textit{\textbf{ViT-B/16 backbone (pretrained on ImageNet-21K)}}} \\
\midrule
\multirow{3}{*}{NFL+LoRA} & \multirow{3}{*}{CIL} & CIFAR-100 & $lr$: 1e-4, $r$: 8, $\lambda_{\text{F}}$: 1.0, $p$: 2.0, $\alpha$: 0.5, $\gamma$: 0.95 \\
 &  & ImageNet-R & $lr$: 5e-5, $r$: 8, $\lambda_{\text{F}}$: 1.0, $p$: 2.0, $\alpha$: 0.5, $\gamma$: 0.95 \\
 &  & ImageNet-A & $lr$: 5e-5, $r$: 8, $\lambda_{\text{F}}$: 0.5, $p$: 2.0, $\alpha$: 0.5, $\gamma$: 0.95 \\
\midrule
\multirow{3}{*}{CL-LoRA} & \multirow{3}{*}{CIL} & CIFAR-100 & $lr$: 1e-4, $r$: 8 \\
 &  & ImageNet-R & $lr$: 5e-5, $r$: 8 \\
 &  & ImageNet-A & $lr$: 5e-5, $r$: 8 \\
\midrule
\multirow{3}{*}{EWC-LoRA} & \multirow{3}{*}{CIL} & CIFAR-100 & $lr$: 1e-4, $r$: 8, $\lambda$: 25, $\gamma$: 1.0 \\
 &  & ImageNet-R & $lr$: 5e-5, $r$: 8, $\lambda$: 50, $\gamma$: 0.95 \\
 &  & ImageNet-A & $lr$: 5e-5, $r$: 8, $\lambda$: 25, $\gamma$: 0.95 \\
\bottomrule
\end{tabular}
\end{table*}
\end{document}